\documentclass[times,review,10pt,3p]{elsarticle}
\usepackage{algorithm}
\usepackage{algorithmic}
\usepackage{subcaption} 
\usepackage{graphicx}
\usepackage{amsmath}
\usepackage{graphicx}
\usepackage{booktabs}
\usepackage[table]{xcolor}
\usepackage{amssymb}
\usepackage{newfloat}
\usepackage{listings}
\usepackage{wrapfig}
\usepackage{hyperref}
\usepackage{multirow}    
\usepackage[table]{xcolor} 
\usepackage{array}       

\journal{Pattern Recognition}
\begin{document}

\begin{frontmatter}



\title{DecoderTracker: Decoder-Only End-To-End method for Multiple-Object Tracking} 



\author[a]{Pan Liao}
\ead{liaopan@mail.nwpu.edu.cn} 

\author[a]{Feng Yang*}

\ead{yangfeng@nwpu.edu.cn} 

\author[a]{Di Wu }
\ead{wu_di821@mail.nwpu.edu.cn} 

\author[a]{Wenhui Zhao}
\ead{zwh2024202513@mail.nwpu.edu.cn} 

\author[a]{Jinwen Yu}
\ead{yujinwen@mail.nwpu.edu.cn} 

\author[a]{Dingwen Zhang}
\ead{zdw2006yyy@nwpu.edu.cn} 
\address[a]{School of Automation, Northwestern Polytechnical University, Xi'an, 710072, Shaanxi, China}

\tnotetext[1]{This research did not receive any specific grant from funding agencies in the public, commercial, or not-for-profit sectors.}
\cortext[1]{Corresponding author} 

\begin{abstract}
Decoder-only methods, such as GPT, have demonstrated superior performance in many areas compared to traditional encoder-decoder structure transformer methods. Over the years, end-to-end methods based on the traditional transformer structure, like MOTR, have achieved remarkable performance in multi-object tracking. \textcolor{black}{However,the} substantial computational resource consumption of these methods, coupled with the optimization challenges posed by dynamic data, results in less favorable inference speeds and training times. To address the aforementioned issues, this paper optimized the network architecture and proposed an effective training strategy to mitigate the problem of prolonged training times, thereby developing \textcolor{black}{DecoderTracker}, a novel end-to-end tracking method. Subsequently, to tackle the optimization challenges arising from dynamic data, this paper introduced DecoderTracker+ by incorporating a Fixed-Size Query Memory and refining certain attention layers. Our methods, without any bells and whistles, outperforms MOTR on multiple benchmarks, \textcolor{black}{featuring a 2 to 3 times faster inference than MOTR}, respectively. The proposed method is implemented in open-source code, accessible at \href{https://github.com/liaopan-lp/MO-YOLO}{https://github.com/liaopan-lp/MO-YOLO}.
\end{abstract}




\end{frontmatter}



\section{Introduction}

Multi-Object Tracking (MOT) is a critical task in the field of video analysis, primarily aimed at inferring and predicting the motion trajectories of objects or instances across a sequence of continuous images. The success of transformer \cite{vaswani2017attention} methods in various domains in recent years is well documented, and transformer-based end-to-end MOT methods such as MOTR \cite{zeng2022motr} and MOTIP \cite{gao2025multiple} have recently been introduced in the MOT domain. These methods offer advantages over the widely used tracking-by-detecting (TBD) methods \cite{bewleySimpleOnlineRealtime2016a,zheng2024motion,maggiolinoDeepOCSORTMultiPedestrian2023,shim2024confidence}, such as eliminating the need for manual design or feature selection.  Among them, the MOTR series based on the Tracking-by-Query(TBQ) paradigm extends object queries to track queries and achieves tracking by propagating these track queries across frames. This category of methods exhibits exceptional extensibility, with their core ideas frequently adopted by models in multiple domains such as 3D detection\cite{lin2023sparsev34d}, motion prediction\cite{shi2022motion}, and video understanding\cite{zhong2023stoa}. In contrast, MOTIP\cite{gao2025multiple}, another end-to-end MOT method currently achieving higher accuracy, is somewhat inferior in this aspect. Its tracking mechanism is more akin to the TBD approach, except that its detection and tracking networks are jointly optimized during training. Thus, in-depth research on TBQ methods will not only significantly advance the development of MOT technology but also hold important implications for driving MOT toward greater efficiency and universality.

\begin{wrapfigure}{r}{0.5\textwidth}

\begin{minipage}{0.5\textwidth}
    \centering
	\includegraphics[width=1\textwidth]{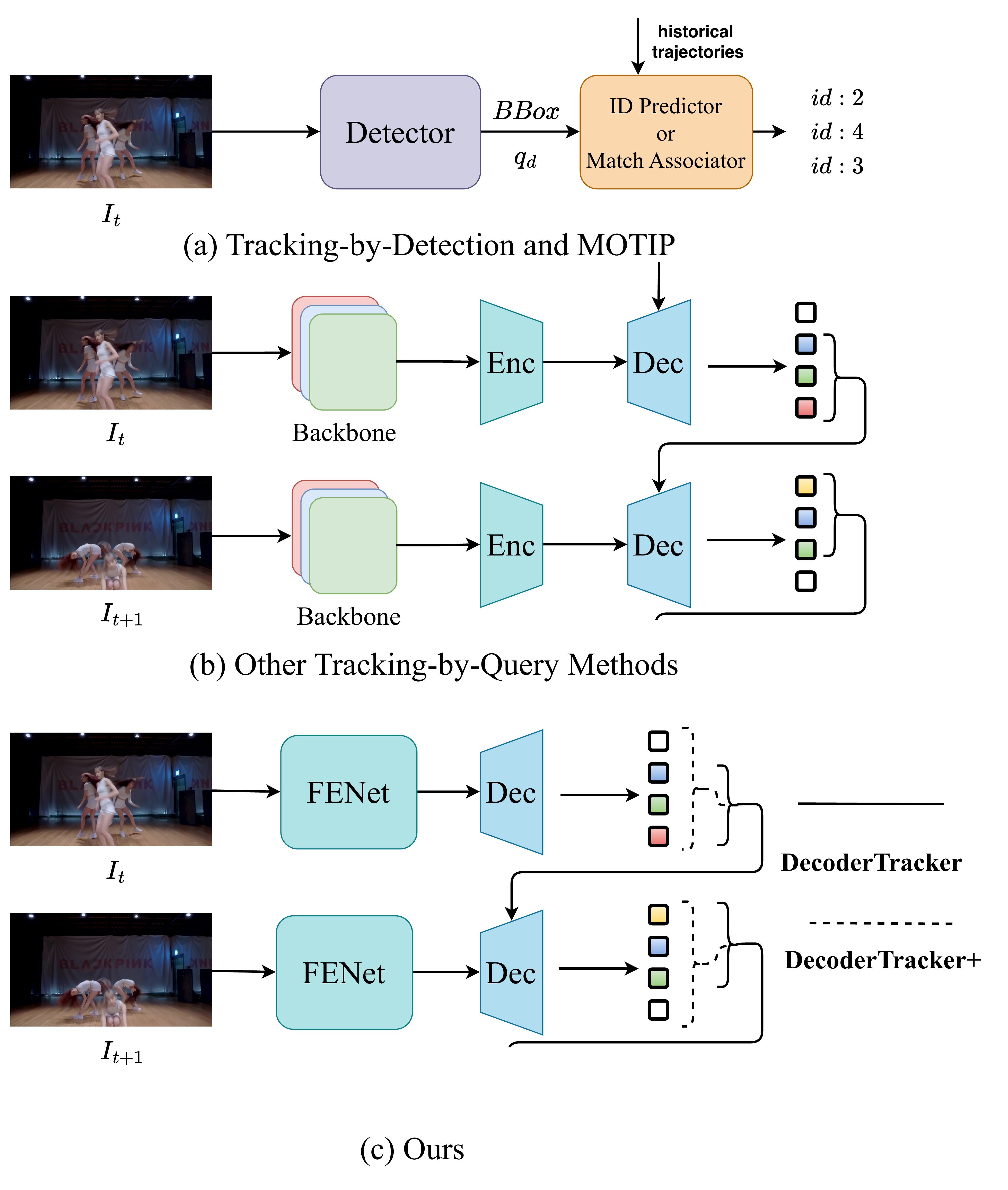}
	\caption{
        \color{black}{Illustration of diverse MOT pipelines. For MOTIP and TBD, a detector first yields detections, followed by ID assignment networks or matching associators for tracking. In contrast, MOTR-series tracking-by-query methods use track queries to model targets and propagate them frame-by-frame. Our approach also employs tracking-by-query but features a simpler network; the proposed FSQM overcomes dynamic data challenges inherent in other tracking-by-query methods.}
	}
    \label{related_work}

\end{minipage}
\end{wrapfigure}

However, the inference speed of these TBQ methods is often suboptimal. Their RNN-like linear inference structure makes them less GPU-friendly during training, and this linear training paradigm also renders data augmentation techniques commonly used in detection tasks inapplicable. \color{black}{Upon in-depth analysis, it becomes clear that the core mechanism of TBQ is accomplished by the decoder, modeling object trajectories and temporal correlations through query propagation and attention interactions, while the encoder merely handles static image feature extraction and contributes little to the core logic of dynamic tracking.}

\color{black}This characteristic is closely aligned with the successful application of decoder-only architectures in sequence modeling. For instance, large models such as GPT\cite{brown2020language}, which utilize decoder-only structures, have demonstrated exceptional performance in complex sequence tasks. Their advantages lie in efficiently modeling long-range dependencies via multi-head attention while avoiding redundant computations from encoder-decoder interactions. In Deformable-DETR \cite{zhu2020deformable} which is the baesmodel of most end to end MOT method training , we observed an imbalance in the learning of image features and object features, with the learning of object features lagging significantly. This is closely related to the high computational cost yet low contribution of the encoder: As noted in \cite{lin2022d}, the encoder in Deformable-DETR accounts for 49\(\%\) of GFLOP but contributes only 11\% to AP, prompting us to question whether MOTR faces the same problem. Accordingly, we draw on the efficiency advantages of decoder-only architectures in sequence modeling and the design principles of real-time detection networks to construct a decoder-only tracking network. By removing the redundant encoder and concentrating computational resources on the decoder, which is critical for tracking, we lay the foundation for improving the inference speed and training efficiency of the end-to-end MOT. Subsequent experiments have also confirmed that our decoder-only model achieves a certain degree of performance improvement over the encoder-decoder-based MOTR.



\color{black}{}
At the same time, we discovered that the original training process for MOTR is one of strong supervision, which led us to question whether using relatively weaker supervision and allowing the method to engage in less stringent learning might also be effective. Is there a possibility that this process could yield unexpected benefits? It is important to note that weak supervision, or self-supervised learning, has achieved significant success in many computer vision tasks \cite{chen2020simple,chen2020big}. Based on this idea, we proposed the Tracking Box Selection Process (TBSP) which is mainly utilized \textcolor{black}{during training}. Thus, \textcolor{black}{we propose DecoderTracker, a highly efficient end-to-end TBQ tracking network. Its fundamental departure from prior approaches is visually discernible in Figure \ref{related_work}.} 

After completing the aforementioned architectural optimizations mentioned above, we observed that while the training duration was significantly optimized, the inference speed did not improve as expected. This phenomenon prompted us to delve deeper into the issue. In theory, streamlining the network architecture should more notably enhance inference speed rather than training efficiency. To investigate this anomaly, we systematically analyzed the time consumption distribution across various modules. The results revealed that the decoder module accounted for the majority of the time, consuming nearly two-thirds of the model's total inference time, which is evidently unreasonable. Through further cross-model comparative analysis, we identified that this issue also exists in MOTR but is absent in its base model, Deformable DETR. Controlled variable experiments indicated that fixing the number of queries significantly reduces decoder latency. Considering factors such as GPU memory usage, we ultimately pinpointed the root cause to be the dynamic query mechanism. This mechanism introduces unnecessary overhead during inference, such as inefficient GPU memory allocation in modern deep learning frameworks ,e.g., PyTorch/TensorFlow, and may also hinder compilation optimizations like operator fusion due to dynamic computation graphs. To address this issue, we proposed the \textbf{Fixed-Size Query Memory (FSQM)} for handling fixed-query tracking. By integrating FSQM and optimizing the attention layers in the original decoder and TAN modules, we introduced DecoderTracker+.  Experimental results demonstrated that, although DecoderTracker+ theoretically increases computational load, its inference speed is significantly faster than the original version (from 19.6 fps to 28.8 fps) due to static data processing, while maintaining comparable tracking performance.


In summary, the contributions of this paper are as follows:

\begin{itemize}
	\item [1)]
	The primary contribution involves the development of a novel end-to-end tracking network, DecoderTracker, which is a decoder-only method. Compared to MOTR, this network achieves faster inference speeds and improved tracking performance.
	\item [2)]
	The second key innovation involves the introduction of a unique training strategy characterized by a three-stage process. At the same time, we incorporated TBSP, a weakly supervised training strategy designed for preliminary training of end-to-end MOT methods. \color{black} These strategies accelerate method convergence, reduce overall training time by enhancing the efficiency of the training process.
	
	\color{black}
	\item [3)]
	
	Finally, by leveraging FSQM and optimizing certain self-attention layers within the network, we developed DecoderTracker+, which addresses the additional latency caused by dynamic data. This provides a fully viable technical pathway for the engineering deployment of MOTR-like end-to-end MOT models.

\end{itemize}

\section{Related work}

\subsection{MOTR Series}

Unlike tracking-by-detection methods exemplified by SORT \cite{bewleySimpleOnlineRealtime2016a,zhangByteTrackMultiobjectTracking2022}, and previous transformer-based MOT methods \cite{sun2020transtrack,meinhardt2022trackformer}, MOTR \cite{zeng2022motr} is the first method to achieve end-to-end multi-object tracking. It extends the object query from DETR \cite{carion2020end} into a track query, thereby enabling tracking capabilities. It uses a learnable positional embedding and multi-scale deformable attention to predict object positions and classes in frames, eliminating the need for explicit data association or post-processing. Key innovations include the Track Query Aggregation with Learned Attention (TALA), which uses a query memory bank and multi-head attention for methoding long-term temporal relationships, and the introduction of Collective Average Loss (CAL) and Temporal Aggregation Network (TAN) to resolve target conflicts and improve temporal information integration. These features establish MOTR as a significant advancement in multi-object tracking technology.

%
%

MOTRv2 \cite{zhang2023motrv2} enhances the detection capabilities of the original MOTR by integrating the pre-trained object detector YOLOX \cite{ge2021yolox}. This integration involves using YOLOX generated proposal boxes as anchors to initialize object queries, a method that not only bolsters detection accuracy but also helps to alleviate conflicts between detection and association tasks, all the while maintaining the query propagation features inherent to MOTR.

Similarly, MeMOTR \cite{gao2023memotr} innovates by incorporating track queries that predict target trajectories iteratively across video frames. This approach utilizes a tracklet-aware label assignment method to ensure a one-to-one correspondence between track queries and actual target trajectories, addressing challenges with newly introduced object queries.

MOTRv3 \cite{yu2023motrv3} and CO-MOT \cite{yanBridgingGapEndtoend} focus on optimizing query interactions and label assignment within the MOTR framework. MOTRv3 adopts a release-fetch supervision strategy for balanced label assignments, while CO-MOT employs cooperative label assignment with "shadows" to enhance query diversity. DNMOT\cite{fu2023denoising} addresses severe occlusions through denoising training (simulating occlusion with noise) and a Cascaded Mask Module that coordinates query interactions to prevent suppression of occluded trajectories. DQFormer\cite{10.1145/3735510}, meanwhile, uses Decoupled Query Augmentation to separate detection and association query branches, reducing feature conflicts. A

\subsection{Tracking by Detection}

\textcolor{black}{Tracking by Detection frameworks are among the most classical and widely used approaches for MOT.} These methods typically employ a detector to first identify objects and then use various other methods for tracking.

In past years, the main methods within the TBD category were SORT \cite{bewleySimpleOnlineRealtime2016a} and its derivative methods. The primary approach of these methods is to use a Kalman Filter to predict the positions of targets in the next frame and then perform data association with the detection results provided by the detector. Many subsequent improvements to these methods \cite{maggiolinoDeepOCSORTMultiPedestrian2023,zhangByteTrackMultiobjectTracking2022,shim2024confidence} have introduced re-identification (reid) networks, incorporating object appearance features as an additional factor for association. There are also a series of methods like OC-SORT \cite{cao2023observation} and MotionTrack\cite{zheng2024motion} that purely improve the Kalman Filter-based tracking method using kinematic methods. In recent years, within the TBD paradigm, a new wave of methods has emerged, such as SportMamba\cite{khanna2025sportmamba}, TWiX\cite{miah2025learning}, MSPNet\cite{van2024visual} and Glan\cite{liu2024glan}. These approaches leverage advanced neural network architectures—including spatiotemporal modeling networks and query-based attention mechanisms—to refine the post-detection tracking process, moving beyond the traditional reliance on Kalman filtering or standalone re-identification modules. Unlike earlier methods that primarily relied on kinematic predictions or appearance features for association, these newer frameworks integrate deeper spatiotemporal reasoning into the tracking pipeline, enabling more robust handling of complex scenarios such as long-term occlusion and fast motion.

\begin{figure*}
	\centering
	\includegraphics[width=1\textwidth]{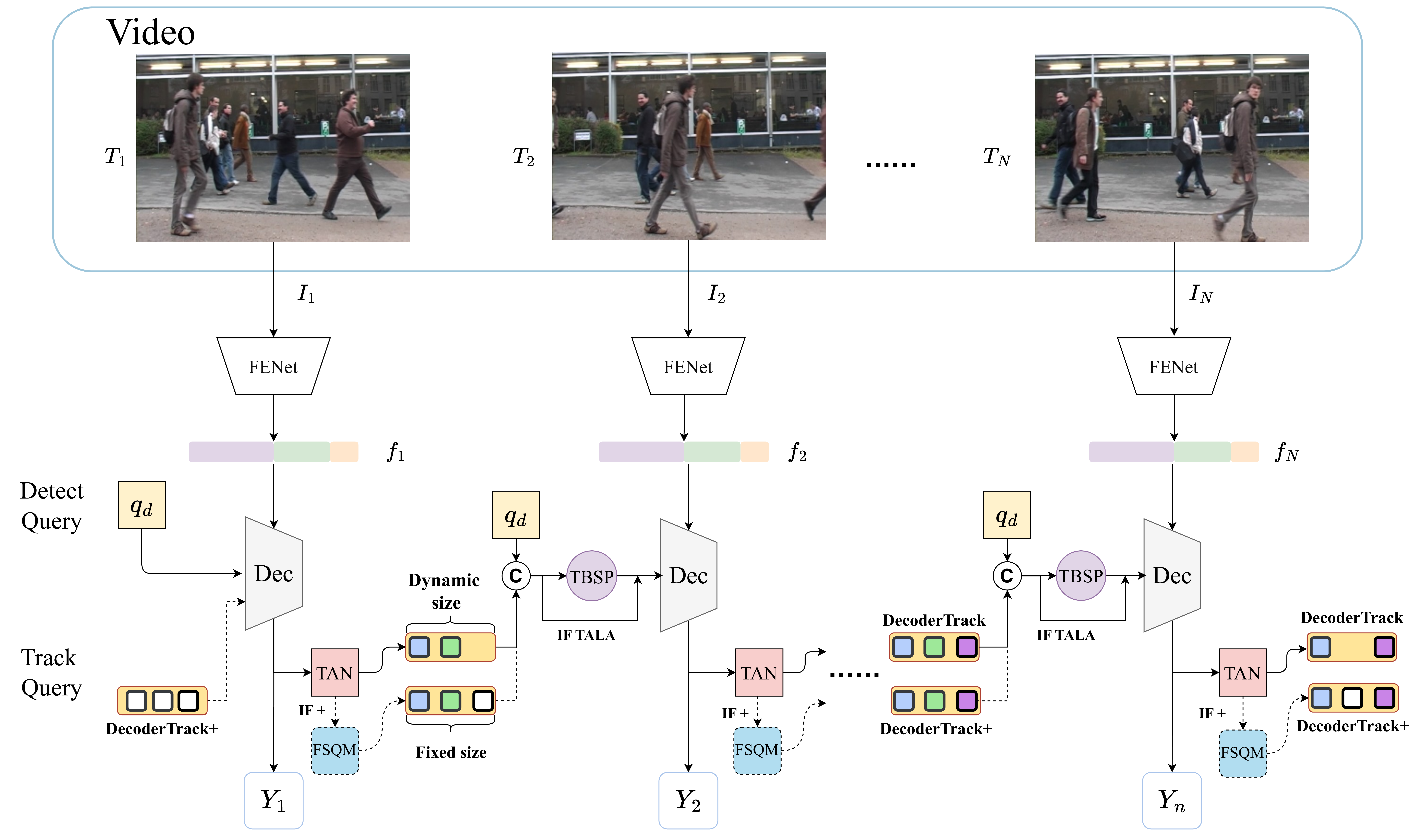}
	\caption{The overall structure of DecoderTracker. By merging the detect queries ($q_d$) and track queries ($q_{tr}$), the decoder (Dec) processes them to generate hidden states. To put it simply, $q_{tr}$ represents the objects in the video sequence that need to be tracked continuously, while $q_d$ represents the new objects appearing in the video frame. When the method is trained using the TALA strategy, the TBSP can be \textbf{skipped}. The process of DecoderTracker+ is consistent with this scenario, but the main difference lies in the updating of $q_{tr}$ and whether it is of fixed length. The FSQM of DecoderTracker+ pads $q_{tr}$ with all-zero queries. This approach effectively addresses optimization issues caused by dynamic data.}
	\label{famework}
\end{figure*}

\subsection{Real Time Detection Method}



When discussing real-time detection methods, most people initially think of YOLO \cite{redmon2016you}, which transformed the problem of object detection into a regression method, achieving efficient object detection. The method divides an image into grids, with each grid responsible for predicting the bounding boxes and class information of objects within its area, treating the task as a regression problem of coordinates and dimensions. \textcolor{black}{Among recent variants in the YOLO family, YOLOv8 and its upgraded version YOLO11\cite{yolov8_ultralytics} stand out for their superior stability and migration capability, making them the most reliable choices for practical deployment.}

In recent years, \textcolor{black}{a real-time detection method based on DETR, known as RT-DETR \cite{zhao2024detrs}, has also shown extraordinary performance. RT-DETR has also demonstrated extraordinary performance, achieving optimized inference speeds by designing an efficient encoder and simplifying its network architecture. In particular, YOLO11 primarily focuses on modifications in the head section, while its core feature extraction framework remains largely consistent with YOLOv8. Given the proven stability and mature migration performance of YOLOv8 in various scenarios, our work thus draws more on YOLOv8 for FENet design, ensuring reliability in our MOT pipeline, which prioritizes low latency over the transformer encoder.}

\section{Method}
\subsection{DecoderTracker}


\color{black}
\subsubsection{Architecture of Proposed Methods}

The overall architecture of DecoderTracker and DecoderTracker+ are illustrated in the provided Figure \ref{famework}. \textcolor{black}{Each frame of the input video is fed into the \textbf{F}eature \textbf{E}xtraction \textbf{Net}work(FENet) to obtain multi-scale feature maps.} For the first frame, where there is no tracking information available, a fixed-length learnable detect query (referred to as $q_d$ in the figure) is \textcolor{black}{inputted} to the decoder. For subsequent frames in the video sequence, the tracking queries from the previous frame and the learnable detect queries are concatenated and undergo filtering before being input to the decoder. These queries interact with the image features in the decoder to generate hidden states for bounding box predictions. \textcolor{black}{Additionally, MOTR’s Track Association Network (TAN) is employed to update the track queries, providing trajectory queries for the next frame.}



\subsubsection{Design of FENet}
\begin{figure*}
	\centering
	\includegraphics[width=1\textwidth]{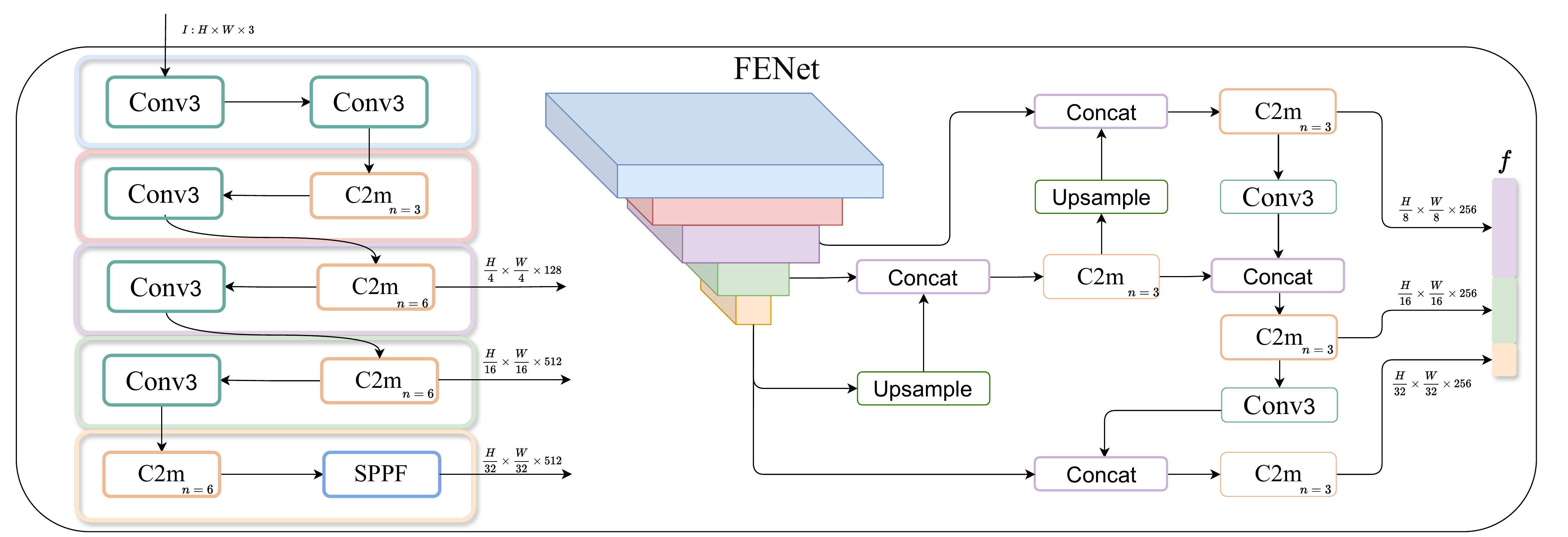}
	\caption{
        \color{black}The structure of FENet. It bears certain similarities to the structure of YOLOv8's backbone and neck, yet there are distinct differences. The left part of the figure corresponds to the backbone, which consists of C2m and Conv3 modules (detailed in Figure \ref{method_detail}) to extract initial multi-scale features from input images. The right part represents the neck, where features from the backbone are processed through upsampling and concatenation operations—specifically, high-resolution features from upper layers are fused with low-resolution features from lower layers via C2m modules—to generate the final three-scale feature maps. These feature maps are then fed into the decoder and introduced to the cross-attention module to interact with the queries. Similar to the Backbone and Encoder in the MOTR series, FENet focuses solely on extracting features from the current video frame, while the implicit spatio-temporal feature extraction is managed by the Decoder.
        }
	\label{detail}
\end{figure*}

\textcolor{black}{FENet is a lightweight backbone-plus-FPN network derived from YOLOv8. It replaces the ResNet-50 backbone and the multi-layer transformer encoder in MOTR, delivering three-scale feature maps to the decoder.} The reason for not adopting the latest YOLOv9 or YOLOv10 is due to its slower training speed compared to YOLOv8. Through FENet, we encode an image of dimensions $ H \times W \times 3$ into three tensors of sizes $\frac{H}{8} \times \frac{W}{8} \times256$,$\frac{H}{16} \times \frac{W}{16} \times256$, $\frac{H}{32} \times \frac{W}{32} \times256$, which are then fed into the decoder along with the track and detect queries. 

\textcolor{black}{Our FENet} is intended to enhance the efficiency of the method and increase its inference speed. \textcolor{black}{Thus, our design primarily resembles the YOLOv8 network structure. The only differences are in the neck structure, which we adapt to fit the input requirements of the decoder.}

\begin{figure*}
	\centering
	\includegraphics[width=1\textwidth]{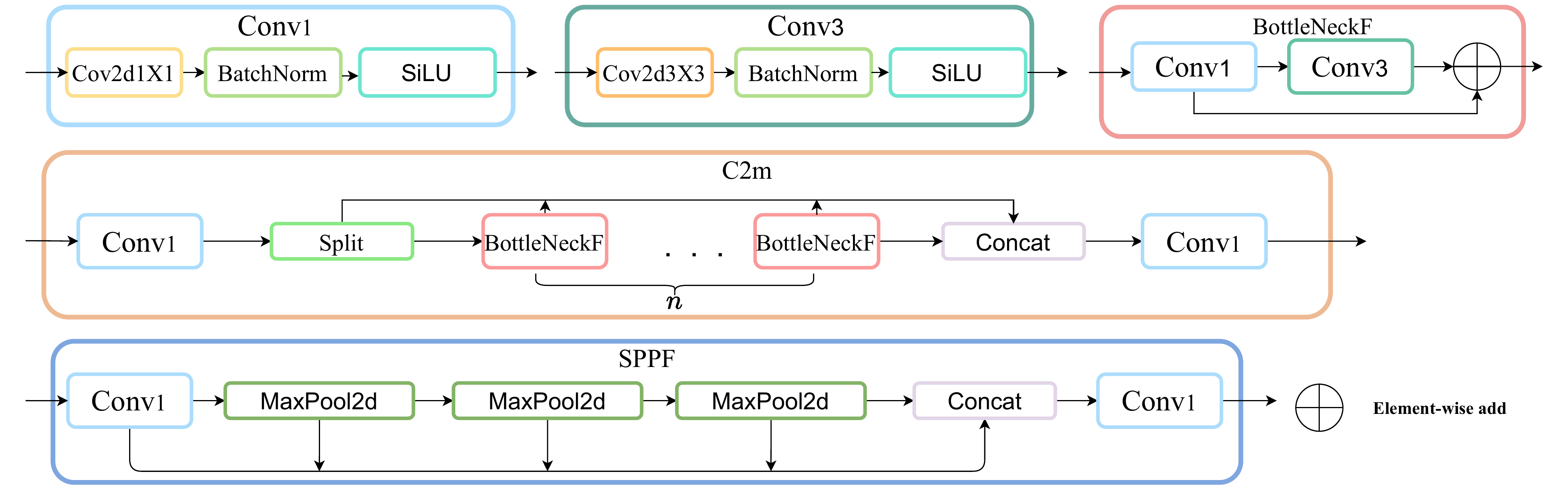}
	\caption{
\color{black}
    Specific module details. Modules not specified in the diagram are network structures or APIs that come with PyTorch. The C2m module, composed of Split, BottleNeckF, and Concat operations, is a core component in both the backbone and neck of FENet (as shown in Figure \ref{detail}), responsible for feature refinement and propagation. The SPPF module, located in the backbone, uses multiple MaxPool2d layers to enhance multi-scale feature aggregation. These modules are primarily constructed based on CNN, capable of efficiently extracting image features. The employment of residual connections ensures that issues like gradient explosion and vanishing gradients are largely avoided.
	}
	\label{method_detail}
\end{figure*}

\color{black}

\subsubsection{Tracking Box Selection Process}
\label{TBSP}


\begin{algorithm}[H]
	\caption{TBSP flow}
	\label{trick_formulae}
	\begin{algorithmic}[1]
		\REQUIRE $\hat{\omega}_{tr}^{t-1}$: Track queries assignment, $\hat{\omega}_{det}^{t-1}$: Detect queries assignment, $\delta_{iou}$: IOU threshold
		\STATE $[BBox_1^{tr}, BBox_2^{tr}, \dots, BBox_n^{tr}] \leftarrow$ BBoxes from $\hat{\omega}_{tr}^{t-1}$
		\STATE $[BBox_1^{det}, BBox_2^{det}, \dots, BBox_m^{det}] \leftarrow$ BBoxes from $\hat{\omega}_{det}^{t-1}$
		\FOR{ $BBox_j^{det}$ in $[BBox_1^{det}, \dots, BBox_m^{det}]$}
		\FOR{ $BBox_i^{tr}$ in $[BBox_1^{tr}, \dots, BBox_n^{tr}]$}
		\STATE Compute $IOU(BBox_j^{det}, BBox_i^{tr})$
		\IF{$IOU > \delta_{iou}$}
		\STATE Discard $BBox_j^{det}$
		\STATE \textbf{break}
		\ENDIF
		\ENDFOR
		\ENDFOR
		\STATE $\omega_{\det_f}^{t-1} \leftarrow$ Remaining detect queries assignment
	\end{algorithmic}
\end{algorithm}

The initial design of the TBSP was intended merely to equip a detection network with basic tracking functionality. We observed that when a detection network directly employs the MOTR \cite{zeng2022motr} method to update track queries, the tracking boxes for the same object tend to increase over time. This accumulation was found to be due to each newly generated detect query invariably being assigned to an already tracked object. In the DETR, a detect query can be assigned to any object in the image, as label assignment is determined through bipartite matching between all detection queries and ground truth data. MOTR addresses this by utilizing tracklet-aware label assignment (TALA), which suppressively modulates the generation of detect queries similar to track queries during training, thus mitigating this issue.

Consequently, we designed a simple filtering mechanism, TBSP, which enables the detection network to possess some tracking capabilities without additional training. However, given the network's inherently weak tracking ability and the lack of robustness in TBSP, the tracking performance at this stage is quite limited. Nevertheless, we also discovered that using TBSP as a preprocessing step for TALA, and training the network in this fashion, can lead to some unexpected benefits. Compared to TALA, TBSP can be considered a form of self-supervised or weak supervision during training. This is because TBSP does not forcibly assign a specific query to correspond to an object; instead, it allows the method to optimize itself.

Formally, we denote the predictions of track queries as $\widehat{Y}_{t r}$ and predictions of detect as $\widehat{Y}_{\text {det }} . Y_{\text {new }}$ is the ground-truths of newborn objects. The label assignment results for track queries and detect queries can be written as $\omega_{tr}$ and $\omega_{\text{det}}$. For frame $i$, label assignment for detect queries is obtained from bipartite matching among detect queries and newborn objects. We know that both track query and detect query contain information about the object's bounding box (BBox). Let's denote the BBox in $\hat{\omega}_{tr}^{t-1}$ as $[BBox _{1}^{tr}$, $BBox _{2}^{tr}, ..., BBox _{n}^{tr}]$, where $n$ is the number of track queries at time $t-1$. Each query assignment in $\hat{\omega}_{det}^{t-1}$ is denoted as $[\omega _{1}^{det}, \omega _{2}^{det}, ..., \omega _{m}^{det}]$, and its corresponding BBox is denoted as $[BBox _{1}^{det}, BBox _{2}^{det}, ..., BBox _{m}^{det}]$.

For each element in $[BBox _{1}^{det}, BBox _{2}^{det}, ..., BBox _{m}^{det}]$, calculate the Intersection over Union (IOU) with each element in $[BBox _{1}^{tr}, BBox _{2}^{tr}, ..., BBox _{n}^{tr}]$. If the IOU is greater than $\delta_{iou}$, the BBox corresponding to that detect Query will be discarded. Finally, the remaining detect queries are denoted as $\omega _{\det_f}^{t-1}$.

Similarly, assuming the track query at frame $t$ (where $t > 2$) is denoted as $\omega_{tr}^t$, it can be expressed as:

\begin{align}
	\hat{\omega}_{tr}^{t}\,\,=\omega _{det_f}^{t-1}\cup \,\,\hat{\omega}_{tr}^{t-1}\,\,
\end{align}

The feasibility of TBSP is illustrated in the Algorithm. \ref{trick_formulae}. Incidentally, although TBSP makes DecoderTracker appear not to be a truly end-to-end network, after thorough TALA training, this step can be omitted. Simultaneously, to better illustrate the performance of our method compared to MOTR, we have not adopted the labeling strategy used in CO-MOT \cite{yanBridgingGapEndtoend2023} and MOTRv3 \cite{yu2023motrv3}.

%
%
%
\begin{figure*}
	\centering
	\includegraphics[width=1\textwidth]{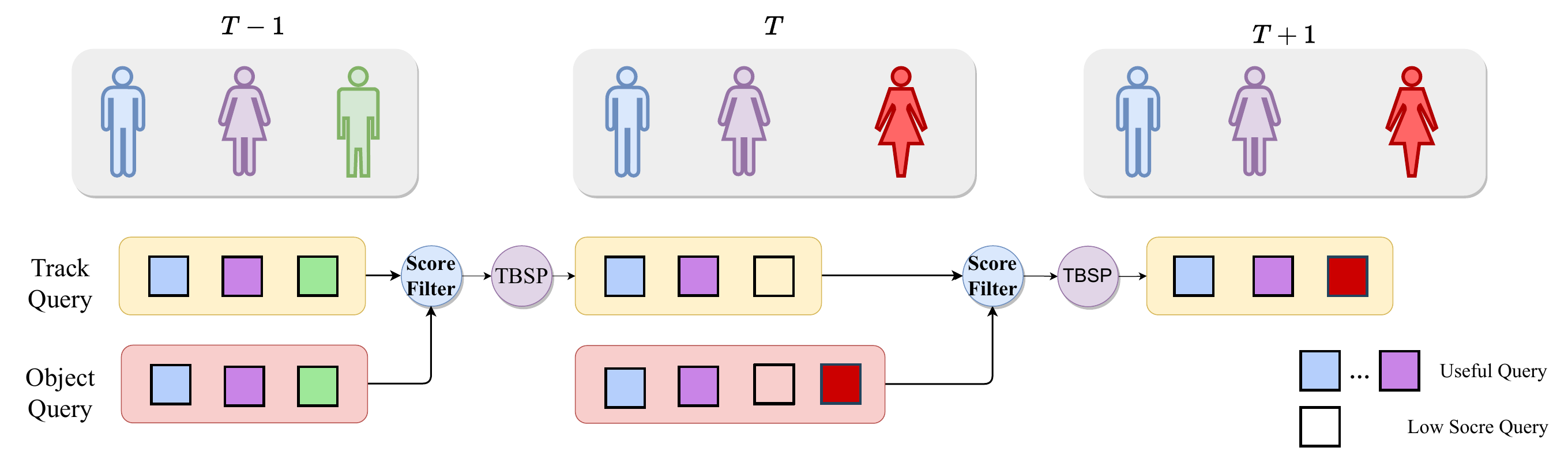}
	\caption{Refinement process for DecoderTracker on partially trained track queries. 
		When DecoderTracker undergoes comprehensive training with the TALA strategy, TBSP will no longer be utilized. At that point, the track query generation method of DecoderTracker aligns with MOTR, as illustrated in the Figure. \ref{related_work}. The process in DecoderTracker+ is similar, but there are some differences in the removal and addition of track queries. The detailed operations are discussed in the subsequent sections.}
	\label{trick}
\end{figure*}%

\subsubsection{Query  and Query Positions}


MOTR was initially built upon Deformable DETR \cite{zhu2020deformable}. The original DETR was considered slow to converge due to its Hungarian matching algorithm. Although Deformable DETR significantly improved the convergence speed of DETR-like methods through architectural enhancements, subsequent works such as DN-DETR \cite{li2022dn} and DAB-DETR \cite{liu2022dab} further corroborated the assertion that the Hungarian matching algorithm was indeed problematic. In DETR and Deformable DETR, the queries are typically initialized as a set of non-interpretable parameters, which is evidently suboptimal for training. Drawing inspiration from RT-DETR's query generation approach, we directly generate queries from the input decoder features \textit{f}. This process can be specifically represented as follows:

\begin{align}
	\begin{aligned}
		q_d &= \text{LayerNorm}(\text{Linear}(f)) \\
	\end{aligned}
\end{align}

One more thing, to expedite method convergence, we implement a Query Positions generation strategy akin to CO-MOT. Drawing inspiration from DAB-DETR \cite{liu2022dab}, we employ a fixed transformation to convert reference points into Query Positions. While experimentation with the RT-DETR approach involving a Multilayer Perceptron (MLP) to transform reference points shows no significant performance discrepancy, it is important to note that the use of MLP substantially increases training time.

\subsubsection{Loss and Training Methods}

To expedite method training, this paper adopts a three-stage training strategy. In the first stage, emphasis is placed on training the network to extract target image features efficiently. The second stage involves training the method on weak tracking aspects using TBSP as a pretraining step for the third stage. Lastly, the third stage employs the TALA strategy to comprehensively enhance the method's tracking performance.

\textbf{In the initial stage}, the network is trained as a detection network, and the loss function is formulated defined as Eq.\ref{eq:total_loss}:

\begin{align}
    \label{eq:total_loss}
    \mathcal{L} \left( \left. \widehat{Y_t} \right|_{\omega_t}, Y_t \right) = 
    \lambda_{\text{cls}} \mathcal{L}_{\text{cls}} + 
    \lambda_{l_1} \mathcal{L}_{l_1} + 
    \lambda_{\text{giou}} \mathcal{L}_{\text{giou}} + 
    \lambda_{\text{enc}} \mathcal{L}_{\text{enc}}
\end{align}
where $\widehat{Y} = \{\widehat{Y}_i\}_{i=1}^N$ denotes frame predictions, $Y = \{Y_i\}_{i=1}^N$ is the ground truth, and $\omega = \{\omega_i\}_{i=1}^N$ represents matched indices. The component losses are defined as follows:

\color{black}{
\begin{align}
    \mathcal{L}_{\text{cls}} 
    &= \sum_{i=1}^{N} -\alpha \left(1 - p_{i,c_i}\right)^\gamma \log\left(p_{i,c_i}\right) 
    \quad  \\
    \mathcal{L}_{l_1} 
    &= \sum_{i=1}^{N} \left\| \hat{b}_i - b_i \right\|_1 
    \quad \text{(L1 loss)} \\
    \mathcal{L}_{\text{giou}} 
    &= \sum_{i=1}^{N} \left[ 1 - \text{GIoU}\left(\hat{b}_i, b_i\right) \right] 
    \quad  \\
    \mathcal{L}_{\text{enc}} 
    &= \sum_{j=1}^{K} \left[ \left\| \hat{b}^{\text{enc}}_j - b_{\sigma(j)} \right\|_1 + \beta \left(1 - \text{GIoU}\left(\hat{b}^{\text{enc}}_j, b_{\sigma(j)}\right)\right) \right] 
    \quad 
\end{align}
Here, $p_{i,c_i}$ denotes the predicted probability for the ground truth class $c_i$ at the matched index $i$. The predicted and ground truth bounding boxes are represented by $\hat{b}_i$ and $b_i$, respectively, each defined by the center coordinates, width, and height. The $j$-th encoder-predicted anchor box is denoted as $\hat{b}^{\text{enc}}_j$. The matching process is governed by $\sigma$, a function that maps the top-$K$ encoder predictions to ground truth instances based on center distance, where $K = \min(\text{top-}K, N)$ is dynamically set to the number of ground truth objects. Hyperparameters include $\alpha=0.25$ and $\gamma=2$ for focal loss and $\beta=1$ for GIoU weight in anchor loss. The loss weights are configured as $\lambda_{\text{cls}}=1$, $\lambda_{l_1}=2$, $\lambda_{\text{giou}}=5$, and $\lambda_{\text{enc}}=5$ to balance classification, L1 regression, GIoU, and encoder losses. In addition, a more specific explanation of these losses can be found in \cite{lin2017focal,rezatofighi2019generalized,liu2022dab}.
}

\color{black}{
In this stage, the network is able to initially learn the appearance characteristics of the objects to be tracked. During this phase, the batch size can be set larger, fully utilizing the computational power of GPUs to reduce training time. Furthermore, various data augmentation techniques \cite{zhang2017mixup,ghiasi2021simple,Wang_2021_CVPR} can be used to expedite the method's extraction of appearance features.} \color{black} Whether these data augmentation methods are applied has a significant impact on the results of this stage, exerting an approximate one point effect on the final tracking accuracy. Since this is not the focus of this paper, detailed results are not enumerated here.

\color{black}

\textbf{The second stage} of training is carried out using the Collective Average Loss of MOTR. During this phase, only MOTR's TAN and our TBSP are utilized.  The loss functions used in this stage and the third stage are as in Eq.\ref{second_loss}.

\begin{align}
	\label{second_loss}
	\begin{aligned}
		&\mathcal{L} \left( \left. \widehat{Y} \right|_{\omega},Y \right)=
		&\frac{\sum_{n=1}^N{\left( \mathcal{L} \left( \left. \widehat{Y}_{tr}^{t} \right|_{\omega _{tr}^{t}},Y_{tr}^{t} \right) +\mathcal{L} \left( \left. \widehat{Y}_{\det}^{t} \right|_{\omega _{\det}^{t}},Y_{\det}^{t} \right) \right)}}{\sum_{n=1}^N{\left( V_t \right)}}, N=5\\
	\end{aligned}
\end{align}
where $V_i=V_{tr}^{i}+V_{det}^{i}$ denotes the total number of ground-truths objects at frame $i$. $V_{tr}^{i}$
and $V_{det}^{i}$ are the numbers of tracked objects and newborn objects at frame i, respectively. $\mathcal{L}$ is the loss of single frame, which is similar to Eq.\ref{eq:total_loss}.

At this stage, the computation of \(\mathcal{L}_{enc}\) also differs from that in the first phase, where the newly generated detection queries are calculated only in conjunction with newly appeared objects. \color{black}Newly appeared objects in the second stage are defined as high-confidence predictions from detect queries (exceeding a confidence threshold) that pass the TBSP filtering process, i.e., their bounding boxes do not significantly overlap (IOU $\leq$ $\delta_{iou}$) with any existing track query (see Section \ref{TBSP} for details on TBSP). This ensures that they are distinct from already tracked objects. \color{black}Moreover, in \(\mathcal{L} \left( \left. \widehat{Y_{tr}^{t}} \right|_{\omega _{tr}^{t}},Y_{tr}^{t} \right)\), \(\mathcal{L}_{enc}\) is absent, and the weight of \(\lambda _{enc}\) is reduced compared to that of the second stage. Thanks to our various strategies and improvements to the network, we achieved huge improvements in training time and inference speed compared to MOTR.

\textbf{The third stage} of training involves the utilization of TALA during the training phase, while simultaneously deactivating the TBSP. Experimental evidence indicates that if the training epochs during this stage are sufficient, TBSP can also be disabled during the inference phase without significantly compromising performance. This finding underscores the robustness of the method, suggesting that once adequately trained in the third stage, TBSP can be turned off during inference with negligible impact on overall performance.


\subsection{DecoderTracker+}
\begin{figure}
	\centering
	\includegraphics[width=\textwidth,height=0.46\textwidth]{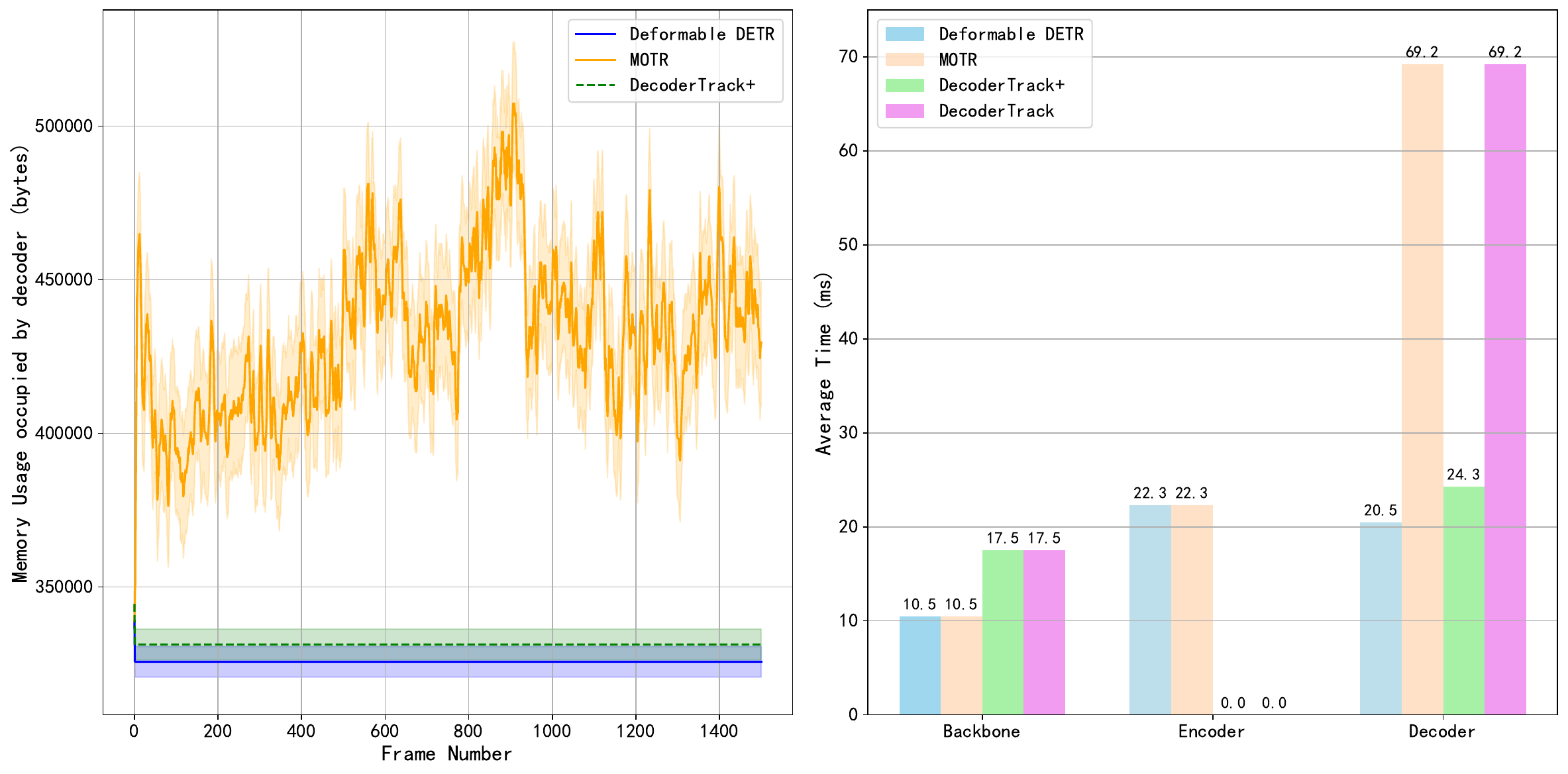}
	\caption{The two figures illustrate the changes in Decoder memory occupancy and the latency of various components for MOTR, its base model Deformable DETR, our DecoderTracker, and DecoderTracker+ on the MOT17-14 sequence. In the right figure, the backbone of our methods is represented by FENet, and since they do not have an Encoder, their latency on the Encoder is 0. The memory changes of DecoderTracker are basically consistent with those of MOTR, so they are not displayed here. It can be clearly seen from the figures that dynamic data leads to unstable memory occupancy of the Decoder and abnormally high latency.}
	\label{memery}
\end{figure}

Following the improvements in previous sections that led to the proposal of DecoderTracker, we observed suboptimal enhancements in inference speed, failing to meet anticipated performance benchmarks. Through latency profiling of individual modules, we identified the decoder as the primary bottleneck limiting acceleration. Subsequent in-depth analyses of GPU memory utilization and query quantity revealed that the core issue likely stems from dynamic data structures. Compared to static data configurations, dynamic data introduces inference challenges such as inefficient memory allocation scheduling, with empirical evidence provided in Figure \ref{memery}.

To address this, we present DecoderTracker+ in this chapter. By implementing a Fixed-Size Query Memory(FSQM) mechanism that maintains a constant number of tracking queries, our approach enables deeper optimization compatibility with modern deep learning frameworks ,e.g., PyTorch, TensorFlow. Furthermore, the deterministic data structure facilitates straightforward conversion to TensorRT, paving the way for accelerated deployment of end-to-end MOT methods on edge devices.

DecoderTracker+ maintains the core architectural framework and operational workflow of its predecessor, as systematically illustrated in Figure \ref{famework}, while introducing strategic innovations primarily in two key dimensions: Fixed-Size Query Memory and Attention Computation Optimization. Subsequent subsections provide detailed technical expositions of these advancements.

\subsubsection{Fixed-Size Query Memory}
FSQM maintains fixed-dimensional queries $q_{tr}^{\mathcal{M}} \in \mathbb{R}^{N_m \times d}$, where $N_m$ denotes the preset maximum number of tracking targets and $d$ represents the feature dimension. Compared to dynamic query mechanisms, this approach introduces:
\begin{equation}
	\hat{\omega}_{tr}^{\mathcal{M}} = (\underbrace{q_{tr}^{\mathcal{M}}}_{\text{static  queries}}, \underbrace{C_{\text{tr}} \in \mathbb{R}^{N_m}}_{\text{confidence vector}},  \underbrace{\mathcal{ID}}_{\text{global ID pool}} , \underbrace{BBox^{\mathcal{M}}}_{\text{bounding box}} )
\end{equation}to model memory for the continuous tracking process. During initialization, set $q_{ftr}^{\mathcal{M}} = \mathbf{0}$ and $C_{\text{tr}}^{(0)} = \mathbf{0}$ for all queries, where all queries are initialized to zero to avoid interference with networks that do not rely on attention mechanisms. Additionally, all inactive queries are also initialized to zero. Moreover, a global ID pool, denoted as $\mathcal{ID}$, is introduced. Each active tracking query is assigned a unique ID that accompanies the query as it moves across video frames. When a query is no longer associated with a tracked object, its corresponding ID is deleted and will not be reused. A new ID is assigned to new active queries. By the way, all inactive queries are assigned an ID of -1 and their bounding boxes are initialized with zero values. \color{black}And the maximum number of tracking targets $N_m$ is set to be equal to the number of detect queries. Theoretically, as long as the actual number of tracked objects does not exceed this preset $N_m$, tracking performance remains unaffected. Based on empirical considerations, we thus set $N_m$ to match the number of object queries. 
\color{black}
During each frame processing, iterative updates are performed as follows:

\begin{enumerate}
	\item \textbf{New query injection}: For detection queries assignment $\{\hat\omega_{\text{det}}^{(i)}\}$ with detection confidence $c_i > \tau_{\text{in}}$, each query is directly assigned the first available ID from the global ID pool (which initially contains the value -1 for new queries). The corresponding query in the memory pool is updated using a momentum update rule:
	\begin{equation}
		\begin{aligned}
			&j = \mathop{\arg\min}\limits_{k} (\mathcal{ID}[k] = -1), \\
			&	\hat{\omega}_{tr}^{\mathcal{M}}{(t)}[j] = \hat{\omega}_{\text{det}}^{(i)} 
		\end{aligned}
	\end{equation}
	
	After updating the query, its corresponding ID is assigned from the global pool.
	
	\item \textbf{Query removal}: For the existing queries in the memory pool, confidence is monitored over time. If a query’s confidence falls below a threshold $\tau_{\text{out}}$ for three consecutive frames, it is considered for removal. Upon removal, the query is reset to zero, and its corresponding ID is set to -1. Specifically:
	\begin{equation}
		\text{If } C_{\text{tr}}^{(t-3)}[n] < \tau_{\text{out}}, C_{\text{tr}}^{(t-2)}[n] < \tau_{\text{out}}, C_{\text{tr}}^{(t-1)}[n] < \tau_{\text{out}},
	\end{equation}
	
	then the query is removed by setting:
	\begin{equation}
		q_{tr}^{\mathcal{M}}[n] = \mathbf{0}, \quad \mathcal{ID}[n] = -1.
	\end{equation}
	
	 The decay ensures that the relevance of older queries decreases over time, while new, active queries retain higher confidence scores.
\end{enumerate}
\color{black}
It is also important to emphasize that, beyond the fixed-size query management (via injection/removal mechanisms) and the unique attention masking strategy of FSQM, the core handling of queries remains consistent with DecoderTracker: including the interaction between detect queries (\(q_d\)) and track queries (\(q_{tr}\)) in the decoder, cross-attention with FENet-derived feature maps, track query updates via TAN, and prediction of bounding boxes and object classes based on query hidden states.

\color{black}
\begin{algorithm}[H]
	\caption{FSQM Online Update Procedure}
	\label{fsqm}
	\begin{algorithmic}[1]
		\REQUIRE New query set $\hat\omega_{\text{det}} = \{q_i, c_i ,BBox_i  \}_{i=1}^M$, memory pool $\hat{\omega}_{tr}^{\mathcal{M}} = (q_{\text{tr}}, C_{\text{tr}}, \mathcal{ID} , BBox_{tr})$
		\ENSURE Updated memory pool $\mathcal{M}'$
		
		\STATE $\hat\omega_{\text{valid}} \leftarrow \{\hat\omega_i | c_i > \tau_{\text{in}} \}$ \COMMENT{Filter valid new queries}
		\IF{$\hat{\omega}_{\text{valid}} \neq \emptyset$}
		\FOR{$q_j \in \mathcal{\hat\omega}_{\text{valid}}$}
		\STATE Assign the first available ID from the global ID pool $\mathcal{ID}$ to $q_j$
		\STATE $k \leftarrow \mathop{\arg\min}\limits_{n} \mathcal{ID}[n] = -1$ \COMMENT{Find the first inactive query}
		\STATE $ q_{\text{tr}}[k] \leftarrow  q_j ,  BBox_{\text{tr}}[k] \leftarrow  BBox_j$ \COMMENT{Momentum update}
		\STATE \textcolor{black}{$C_{\text{tr}}[k] \leftarrow 1$} \COMMENT{Reset confidence to 1.0}
		\ENDFOR
		\ENDIF
		
		\FOR{$n \in \{1,...,N\}$}
		\IF{$C_{\text{tr}}[n] < \tau_{\text{out}}$ for 3 consecutive frames}
		\STATE $q_{\text{tr}}[n] \leftarrow \mathbf{0} , BBox_{\text{tr}}[k] \leftarrow \mathbf{0}$ \COMMENT{Set query and BBox to zero (inactive)}
	\STATE $\mathcal{ID}[n] \leftarrow -1$ \COMMENT{Reset ID to -1}
		\ENDIF
		\ENDFOR
		
		\STATE \textbf{return} $\mathcal{M}' = (Q_{\text{tr}}, C_{\text{tr}}, \mathcal{ID})$
	\end{algorithmic}
\end{algorithm}

\subsubsection{\textcolor{black}{Attention Masking Strategy}}
\label{attention_opt}

Since some queries in FSQM are empty, we need to explore the network structure to ensure that these empty queries do not affect the final results. The network processing these queries consists only of linear layers, normalization layers, dropout, activation function, cross-attention with image features, and finally, self-attention. Our first step is to analyze how empty queries may impact the output of these networks.

\textbf{Firstly, we will analyze some unaffected networks.} These network architectures inherently avoid commingling active and inactive queries in their computational processes. Therefore, during inference, there is no risk of interference arising from their interactions. Our subsequent analysis need only focus on verifying whether these networks produce gradients during the training.

For linear layers, zero-valued queries produce the following output:
\begin{equation}
	W\mathbf{0} + b = b
\end{equation}
Here, $W$ is the weight matrix, $\mathbf{0}$ is the zero-valued query, and $b$ is the bias term. Since the zero-valued query contributes nothing, only the bias remains, which can be neutralized through layer normalization. After applying layer normalization, the output is given by:
\begin{equation}
	\hat{y} = \frac{y - \mu}{\sigma} \cdot \gamma + \beta
\end{equation}
where $y = b$, $\mu$ and $\sigma$ are the mean and standard deviation of the output across the batch, and $\gamma$ and $\beta$ are learnable parameters. Since the bias term $b$ is small in practice, empirical tests show that its $L_2$ norm $\|b\|_2$ is very small, with $\|b\|_2 < 0.03$ in our implementation.

For normalization layers, since the zero-valued queries do not produce any meaningful activation, they do not affect the overall normalization process. The query's zero value is normalized along with other queries, resulting in minimal influence on the normalized output. This can be expressed as:
\begin{equation}
	q_i = \frac{q_i - \mu}{\sigma} \cdot \gamma + \beta \quad \text{for active queries } q_i
\end{equation}
where $\hat{q}_i$ is the normalized output for an active query $q_i$. For inactive queries (those initialized to zero), the effect is negligible due to their zero contribution.

For dropout, which randomly sets a fraction of the inputs to zero during training, inactive queries do not contribute to the dropout process. Inactive queries, having values of zero, remain unaffected by dropout, and the overall output for those queries is still zero. Mathematically, dropout can be expressed as:
\begin{equation}
	\text{Dropout}(\omega_i) = \begin{cases}
		0 & \text{if query is inactive (} q_i = 0 \text{)} \\
		q_i \cdot d & \text{if query is active}
	\end{cases}
\end{equation}
where $d \in [0, 1]$ is a random variable representing the dropout rate. Inactive queries (zero-valued) remain zero regardless of whether dropout is applied.

For activation function, since only GELU is used to process queries, we will only discuss GELU here. In the case of GELU, the output for a zero-valued query is:
\begin{equation}
	\text{GELU}(q_i) = 0.5\omega_i \left( 1 + \tanh\left(\sqrt{\frac{2}{\pi}} (q_i + 0.044715 q_i^3) \right) \right) \quad \text{for } q_i = 0 \implies \text{GELU}(0) = 0.
\end{equation}
Thus, for activation function, zero-valued queries result in zero output, ensuring that they do not influence the subsequent layers.

For cross-attention, given key matrix $K$ and dimension $ d $ of Query, the attention scores become:
\begin{equation}
	\text{Attn}(Q_{\text{zero}}, K) = \frac{\mathbf{0} \cdot K^T}{\sqrt{d}} = \mathbf{0}
\end{equation}
resulting in zero contribution after softmax normalization.

In summary, for linear, normalization, dropout, activation layers and cross-attention, zero-valued queries contribute nothing to the computation, ensuring that they do not interfere with the active queries during the forward and zero-valued in the forward inherently prevent gradient computation, thereby eliminating any impact on the training.

.

\textbf{Regarding the self-attention layers}, since the queries, keys, and values used in computing self-attention are all derived from the queries themselves, inactive queries inevitably introduce interference to the output. While a detailed analysis of this phenomenon is omitted here, we address this issue effectively by implementing a masking mechanism. Here we use the self attention layer in Decoder to illustrate, and the similarity in TAN will not be further elaborated.

When calculating self-attention, we first compute an attention weight matrix $A$:

\begin{equation}
	A = \text{softmax}\left(\frac{q(q')^T}{\sqrt{d}} + \log(\text{Mask}) \right)
\end{equation}

where $q \in \mathbb{R}^{2N \times d}$ is obtained by concatenating $q_{tr}^{\mathcal{M}}$  from FSQM and $q_{det}$, i.e., $q = \text{concat}(q_{tr}^{\mathcal{M}}, q{det})$ in first decoder layer, while in other layers, $ q $ is the output from the previous layer. $(\text{Mask} \in \mathbb{R}^{2N \times 2N}$ is the mask we need to generate. Its dimensions are the same as those of $A$, and it ensures that after applying the softmax function, the unwanted values are set to zero. By appropriately using this mask during the final self-attention output $A \times V$ where $V$ is derived from $Q$ with positional encoding, the effect of empty queries on the final result can be effectively eliminated.

Here, we extend $\mathcal{ID}$ and then generate the mask.
\begin{equation}
	\mathcal{ID}_{\text{ext}} = \text{Concat}(\mathcal{ID}_{\text{tr}}, \mathbf{0}_N)
\end{equation}
where, $\mathbf{0}_N $represents a fixed identifier for detect queries.
\begin{equation}
	\text{Mask}_{ij} = \begin{cases}
		0 & \text{if } \mathcal{ID}_{ext}[i] = -1 \lor \mathcal{ID}_{ext}[j] = -1 \\
		1 & \text{otherwise}
	\end{cases}
\end{equation}
By implementing this mask mechanism, we ensure that values associated with inactive queries in matrix $ A $ are nullified to zero, thereby eliminating their influence on the final output. Notably, as demonstrated through our preceding analyses, the inactive query positions inherently maintain zero values across all decoder layers. Consequently, this masking mechanism uniformly applies to self-attention operations in every decoder layer without requiring layer-specific adaptations.

In terms of the mask in the self-attention layer of the TAN, the generation process and mechanism are largely consistent with those described earlier; however, the dimensions in TAN are fewer, and there is no need to expand $\mathcal{ID}$.

\section{Experiments}

\subsection{Datasets and Metrics}
\textbf{Datasets:} To validate the performance of DecoderTracker and DecoderTracker +, we evaluated the method on three demanding datasets: DanceTrack\cite{sunDanceTrackMultiObjectTracking2022}, MOT17\cite{milanMOT16BenchmarkMultiObject2016}, and KITTI\cite{geigerAreWeReady2012}.
DanceTrack is a vast dataset designed for human tracking, showcasing scenarios with occlusion, frequent crossovers, uniform appearances, and varied body gestures. Comprising 100 videos featuring diverse dance styles, it underscores the significance of motion analysis in multi-object tracking. MOT17 focuses on the tracking of multiple objects in public spaces, primarily by pedestrians. The dataset includes seven scenes, divided into training and testing clips, providing object detection annotations suitable for both online and offline tracking approaches. \textcolor{black}{KITTI covers stereo, flow, odometry, detection and tracking; we use its multi-object tracking split for evaluation.}



\textbf{Evaluation Metrics}: We assessed our method using widely recognized MOT evaluation metrics. The primary metrics employed for evaluation \textcolor{black}{include} HOTA, AssA, DetA ,IDF1 and MOTA\cite{ristani2016performance,luiten2021hota}. These metrics collectively provided a comprehensive and robust evaluation of our method's tracking performance. HOTA assessed the tracking accuracy based on the spatial and temporal overlap between predicted and ground truth tracks. AssA measured the quality of associations between objects, while IDF1 quantified the accuracy of identity tracking. MOTA offered an overall evaluation of tracking accuracy, considering false positives, false negatives, and identity switches. These metrics formed the basis for a thorough evaluation of our method's tracking capabilities.

\subsection{Implementation Details}

The rationale for adopting DecoderTracker originates from our observation of the high training costs and suboptimal inference speed associated with MOTR. Thus, our aim is to propose a new baseline method surpassing MOTR's performance. To ensure a fair comparison, we exclusively employ methods enhancing tracking performance, while omitting improvement techniques utilized in MeMOTR \cite{gao2023memotr}, MOTRv2 \cite{zhang2023motrv2}, MOTRv3 \cite{yu2023motrv3}, and CO-MOT \cite{yanBridgingGapEndtoend2023}.

The majority of experiments were conducted using PyTorch on a single NVIDIA GeForce RTX 4090 GPU. For equitable comparison with MOTR, training on the MOT17 dataset was performed on a single NVIDIA GeForce RTX 2080ti. Additionally, to better contrast with the MOTR series in terms of speed, inference speed tests were carried out on a single NVIDIA V100 GPU. \textcolor{black}{Furthermore}, the image size for the input network is $640\times 640$. The number of object queries is configured as 60 for DanceTrack and KITTI and 300 for MOT17. These values are in line with the previous works \cite{yanBridgingGapEndtoend2023,gao2023memotr}. Consistently, the preset maximum number of tracking targets in the FSQM system is also aligned with these query numbers.


\begin{table*}
	\captionsetup[subtable]{position=top} 
	\begin{subtable}{0.6\textwidth}
		\centering
		\caption{Comparison of Training Time on MOT17}
		\label{device_time}
		\setlength{\tabcolsep}{5pt}
		\begin{tabular}{l c c}
			\hline\noalign{\smallskip}
			Method & Device & Training Time \\
			\noalign{\smallskip}
			\hline
			MOTR\cite{zeng2022motr} & \textbf{8 2080ti} & \textbf{$\sim$
				96.0 hours} \\
			DecoderTracker stage1 & 1 2080ti &  12.1 hours \\
			DecoderTracker stage2 & 1 2080ti &  30.8 hours\\
			DecoderTracker(TBSP) & 1 2080ti &  43.4 hours \\
			DecoderTracker & \textbf{1 2080ti} &  \textbf{68.7 hours} \\
			\hline
		\end{tabular}
	\end{subtable}%
	\begin{subtable}{0.35\textwidth}
		\caption{Comparison of Speed on DanceTrack}
		\label{device_speed}
		\setlength{\tabcolsep}{5pt}
		\begin{tabular}{l c}
			\hline\noalign{\smallskip}
			Method  & FPS \\
			\noalign{\smallskip}
			\hline
			MOTR\cite{zeng2022motr}  & 9.5 \\
			MOTRv2\cite{zhang2023motrv2}  & 6.9 \\
			MOTRv3\cite{yu2023motrv3}  & 10.6 \\
			DecoderTracker(TBSP) & 18.5 \\
			DecoderTracker & \textbf{19.6} \\
			DecoderTracker+ & \textbf{28.8} \\
			\hline
		\end{tabular}
	\end{subtable}
	\caption{Comparison of training time (including all previous stages) and speed between DecoderTracker and MOTR series on a single V100 GPU. The training time of DecoderTracker+ is basically the same as that of DecoderTracker, so it will not be listed separately here.}
	\label{speed_and_fps}
\end{table*}

\begin{figure}
	\centering
	\includegraphics[width=1\textwidth,height=0.5\textwidth]{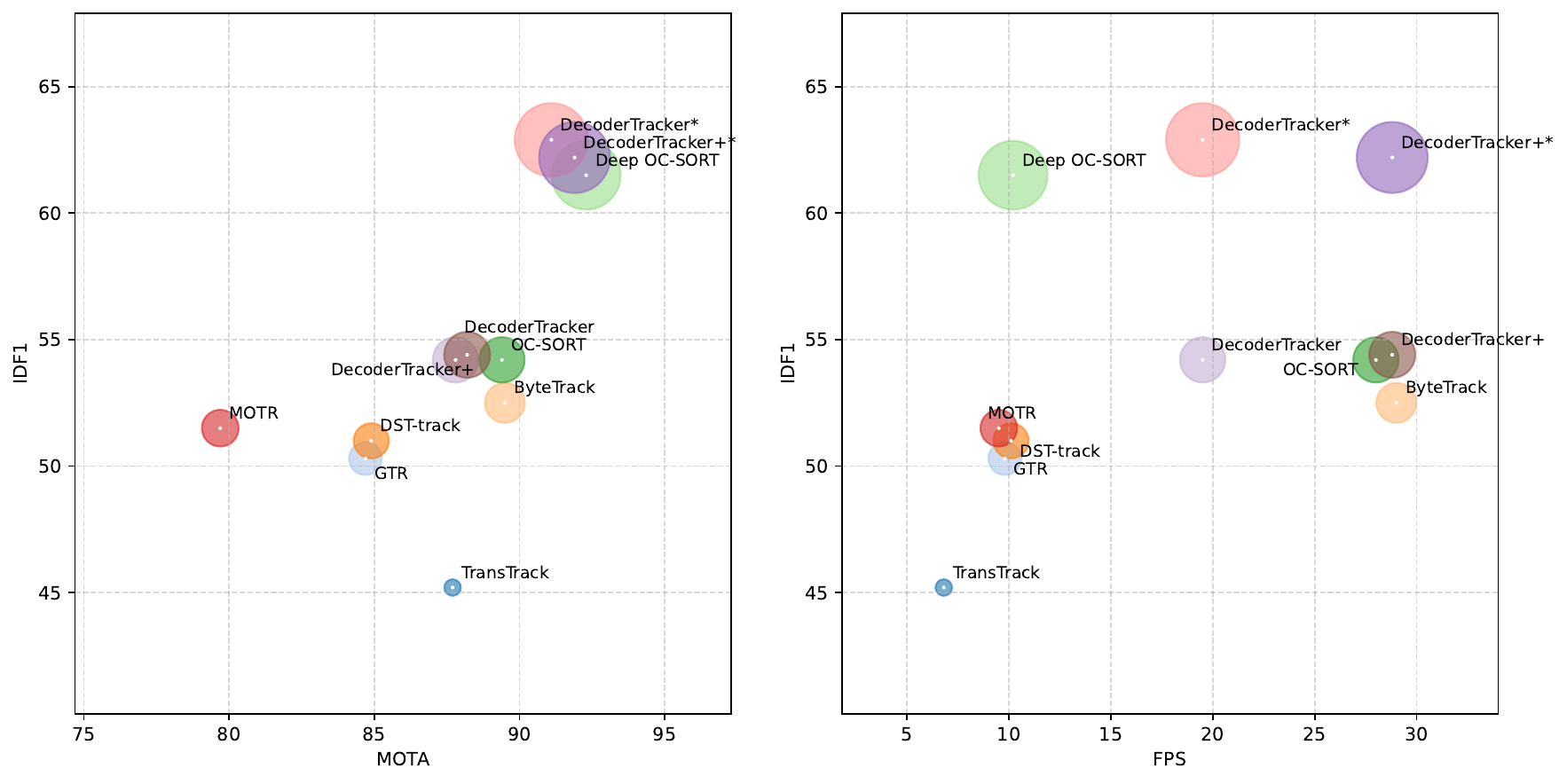}
	\caption{IDF1-MOTA-HOTA and \textcolor{black}{IDF1-FPS-HOTA} comparisons of state-of-the-art trackers with our proposed DecoderTracker and DecoderTracker+ on DanceTrack test sets. For the left image, the horizontal axis is MOTA, the vertical axis is IDF1, and the radius of the circle is HOTA. For the right image, the horizontal axis is FPS, the vertical axis is IDF1, and the radius of the circle is HOTA. ”*” represents additional data was used for training. Our DecoderTracker+* achieves the best IDF1 and HOTA and comparable speed performance.}
	\label{speed_hota_idf1}
\end{figure}

During the initial training, batch sizes were set to 32 on the 4090 and 16 on the 2080ti. Subsequent stages utilized a uniform batch size of 1. \color{black}The random seed was fixed at 42 throughout all experiments to ensure reproducibility.
\color{black}

\textbf{For MOT17}, \color{black}{ following MOTR's data augmentation strategy \cite{zeng2022motr}, we incorporated additional data from CrowdHuman \cite{shao2018crowdhuman}. To convert static images into video-like sequences, we applied random affine transformations (primarily random translation and rotation) to individual images across simulated frames.} \color{black}{Training comprised 120 epochs in the first stage, followed by 30 epochs in the second. The third stage extended to 20 epochs with TBSP, or 55 without. And, to address the issue of the limited scale of the MOT17 dataset, this paper employs a hybrid dataset. It integrates the MOT20 dataset and the complete CrowdHuman dataset, which has been spatially shifted, to expand data diversity. The amount of new data significantly exceeds the benchmark set (by approximately 4.3 times). To balance training bias, a stratified sampling mechanism was designed: within each epoch, samples from the MOT17 dataset are resampled twice, while samples from the supplementary sets are sampled once. This approach not only alleviates data skewness but also enhances the learning of benchmark set features, ensuring that the model maintains its original performance advantages while improving its adaptability to dense scenes.}

\textbf{For DanceTrack}, ensuring parity without extra data like MOTRv2, the first stage was 15 epochs, the second was 9, and the third, if TBSP was used, was 9 epochs, or 18 without TBSP. To better compare with the state-of-the-art (SOTA) methods, when testing on the test set, we also used the CrowdHuman dataset as supplementary data and the validation set of DanceTrack was also used.

\textbf{For KITTI}, initial training involved 80 epochs, followed by 25 in the second stage. The third stage consisted of 30 epochs with TBSP or 70 without. Additionally, approximately 5k images from BDD100k \cite{yu2020bdd100k} were selected for additional training.


\subsection{Main Result}

The comprehensive benchmarking in Table \ref{speed_and_fps} demonstrates DecoderTracker's dual superiority in training efficiency and inference speed over MOTR-series methods. This design achieves 68.7 training hours on a single GPU, equivalent to 7.1× higher GPU-hour efficiency versus MOTR's 768 GPU-hour consumption (8 GPUs × 96 hours). The phased training regimen—12.1 hours for feature embedding, 30.8 hours for temporal modeling, and 43.4 hours for TBSP-enhanced prediction—systematically constructs tracking capabilities while maintaining 83\% shorter wall-clock time than competitors. Inference performance reveals more pronounced advantages: DecoderTracker+ attains 28.8 FPS, 4.1× faster than MOTRv2's 6.9 FPS.

\begin{table*}
    \centering
    \caption{Performance comparison between DecoderTracker and existing methods on the DanceTrack \cite{sunDanceTrackMultiObjectTracking2022} test set. Bold numbers indicate either DecoderTracker results or superior MOTR metrics, while our method uses color highlighting for distinction. Asterisk (*) denotes methods using extra training data.}
    \label{result_dance}
    \begin{tabular}{l *{5}{c}}  
        \toprule
        \multirow{2}{*}{Methods} & \multicolumn{5}{c}{Metrics} \\
        \cmidrule(lr){2-6}  
        & IDF1 & HOTA & AssA & DetA & MOTA \\
        \midrule
        \multicolumn{6}{l}{\textit{Non-End-to-end}} \\  
        \addlinespace[0.2em]  
        TransTrack \cite{sun2020transtrack}         & 45.2 & 45.5 & 27.5 & 75.9 & 87.7 \\
        ByteTrack \cite{zhangByteTrackMultiobjectTracking2022} & 52.5 & 47.3 & 31.3 & 71.6 & 89.5 \\
        OC-SORT \cite{cao2023observation}          & 54.2 & 54.6 & 38.0 & 80.4 & 89.4 \\
        \color{black}MotionTrack \cite{zheng2024motion}          & 58.6 & 58.2 & 41.7 & 81.4 & 91.3 \\ \color{black}
        Deep OC-SORT \cite{maggiolinoDeepOCSORTMultiPedestrian2023} & 61.5 & 61.3 & 45.8 & 82.2 & 92.3 \\
        \midrule
        \multicolumn{6}{l}{\textit{End-to-end}} \\  
        \addlinespace[0.2em]  
        \color{black}
        DNMOT \cite{fu2023denoising}               & 49.7 & 53.5 & --    & --    & 87.7 \\
        \color{black}
        MOTR \cite{zeng2022motr}                   & 51.5 & \textbf{54.2} & \textbf{40.2} & 73.5 & 79.7 \\
        \rowcolor{green!6}  
        DecoderTracker            & \textbf{54.2} & \textbf{54.2} & 35.7 & \textbf{78.7} & \textbf{87.8} \\
        \rowcolor{green!8}  
        DecoderTracker+           & \textbf{54.4} & \textbf{54.5} & 35.6 & \textbf{79.0} & \textbf{88.1} \\
        \rowcolor{green!14}  

        DecoderTracker*           & 62.9 & 60.8 & 47.1 & 81.2 & 91.1 \\
        \rowcolor{green!18}  

        DecoderTracker+*          & 62.2 & 61.2 & 46.0 & 81.8 & 91.9 \\
        \bottomrule
    \end{tabular}
\end{table*}


    As evidenced in Table \ref{result_dance}, DecoderTracker establishes new performance benchmarks for end-to-end multi-object tracking on the challenging DanceTrack benchmark. Our method demonstrates superior detection capabilities, with DecoderTracker and DecoderTracker+ achieving substantially higher DetA scores than all existing end-to-end approaches. This detection advantage directly translates to competitive MOTA performance that approaches state-of-the-art non-end-to-end methods, underscoring the efficacy of our detection-centric architecture. Notably, both variants maintain balanced overall tracking performance comparable to the strongest end-to-end baseline while exhibiting consistent identity preservation metrics (IDF1). The minimal performance variance between DecoderTracker and its enhanced version suggests optimized architectural efficiency rather than fundamental design differences. When augmented with additional training data, our framework achieves transformative performance gains, matching or exceeding leading non-end-to-end methods across multiple metrics. These data-enhanced variants particularly showcase exceptional MOTA and IDF1 results, demonstrating our architecture's scalability and untapped potential. While association accuracy (AssA) remains an area for future improvement relative to specialized baselines, the consistent DetA superiority highlights our method's core strength in target localization. This positions DecoderTracker as a robust foundational framework that bridges the performance gap between end-to-end paradigms and complex multistage trackers while maintaining model efficiency.

\begin{table*}
    \centering
    \caption{Performance comparison between DecoderTracker and existing methods on the MOT17 \cite{milanMOT16BenchmarkMultiObject2016} test set. Bold numbers indicate either DecoderTracker results or superior MOTR metrics, while our method uses color highlighting for distinction. Asterisk (*) denotes methods using extra training data.}
    \label{result_MOT}
    \begin{tabular}{l *{5}{c}}  
        \toprule
        \multirow{2}{*}{Methods} & \multicolumn{5}{c}{Metrics} \\
        \cmidrule(lr){2-6}
        & IDF1 & HOTA & DetA & AssA & MOTA \\
        \midrule
        \multicolumn{6}{l}{\textit{Non-End-to-end}} \\
        \addlinespace[0.2em]
        \color{black} MSPNet \cite{van2024visual} & 74.6 & -- & -- & -- & 71.9 \\ \color{black}
        ByteTrack \cite{zhangByteTrackMultiobjectTracking2022} & 77.2 & 62.8 & 63.8 & 62.2 & 78.9 \\
        OC-SORT \cite{cao2023observation} & 77.5 & 63.2 & 63.2 & 63.2 & 78.0 \\
        \color{black} TWIX \cite{miah2025learning} & 76.2 & 63.1 & 62.1 & 63.1 & 78.1 \\
        \color{black}
        Deep OC-SORT \cite{maggiolinoDeepOCSORTMultiPedestrian2023} & 80.6 & 64.9 & 64.1 & 65.9 & 79.4 \\
        \midrule
        \multicolumn{6}{l}{\textit{End-to-end}} \\
        \addlinespace[0.2em]
        MOTR \cite{zeng2022motr} & 68.9 & \textbf{57.8} & \textbf{60.3} & 55.7 & \textbf{73.4} \\
       \color{black} DNMOT \cite{fu2023denoising} & 68.1 & 58.0 & -- & -- & 75.6 \\\color{black}
       \textcolor{black} {MeMOT} \cite{cai2022memot} & 69.0 & 56.9 & -- & 55.2 & 72.5 \\ 
        \rowcolor{green!6} DecoderTracker & \textbf{70.5} & 55.7 & 55.7 & \textbf{56.0} & 66.5 \\  
        DecoderTracker & \textbf{70.5} & 55.7 & 55.7 & \textbf{56.0} & 66.5 \\
        \rowcolor{green!8}  
        DecoderTracker+ & \textbf{70.4} & 55.8 & 55.2 & \textbf{55.8} & 66.7 \\
       \textcolor{black}{DQFormer} \cite{10.1145/3735510} & 71.6 & 59.2 & 60.6 & 58.1 & 75.2 \\ 
        \rowcolor{green!14}  
        DecoderTracker* & 73.3 & 61.7 & 62.5 & 60.4 & 76.7 \\
        \rowcolor{green!18} 
        DecoderTracker+* & 73.4 & 62.0 & 62.1 & 60.2 & 77.2 \\
        \bottomrule
    \end{tabular}
\end{table*}

In the MOT17 benchmark (Table \ref{result_MOT}), DecoderTracker exhibits strong performance, with a distinct advantage in IDF1 despite marginal deficits relative to MOTR in HOTA and AssA. Its simplified architecture achieves competitive accuracy with reduced training data while maintaining robust overall performance (high MOTA) suitable for real-world applications. Augmented training data further enhances its capabilities: DecoderTracker* and DecoderTracker+* demonstrate significant improvements in IDF1, HOTA, and MOTA, underscoring the positive impact of increased data volume. \color{black}{The slight gaps in HOTA and AssA compared to MOTR can be derived from two factors: first, the streamlined architecture (encoder removed) affects the modeling of long-term temporal correlations, which is critical for HOTA and AssA, which are sensitive to inter-object association accuracy; second, the weak supervision based on TBSP in the three-stage training strategy constrains the precision of the association in complex scenarios, in contrast to the more aggressive TALA strategy of MOTR. To address this, we propose three potential future improvements: integrating the release-fetch supervision from MOTRv3 and the coopetition label assignment strategy from CO-MOT to balance the label assignment weights between detection and tracking queries; enhancing the temporal modeling capability of the TAN module by introducing a dynamic adjustment mechanism for cross-frame attention; and optimizing the matching strategy between static queries and dynamic objects in FSQM to reduce interference from invalid queries on association accuracy. It should be noted that since the focus of this paper is to verify the efficiency advantages of the decoder-only architecture and the effectiveness of the basic training strategy, the improvement directions above will be explored in subsequent studies. Due to the constraints in length and scope of the current research, in-depth exploration of these directions is not included in this paper.} \color{black}

\begin{table*}
    \centering
    \caption{Performance comparison between DecoderTracker and existing methods on the KITTI \cite{geigerAreWeReady2012} test set. Bold numbers indicate the best performance in each category.}
    \label{result_KITTI}
    \begin{tabular}{l *{3}{c} *{3}{c}}  
        \toprule
        & \multicolumn{3}{c}{Pedestrian} & \multicolumn{3}{c}{Car} \\
        \cmidrule(lr){2-4} \cmidrule(lr){5-7}  
        Methods & HOTA & MOTA & AssA & HOTA & MOTA & AssA \\
        \midrule
        \multicolumn{7}{l}{\textit{Non-End-to-end}} \\
        \addlinespace[0.2em]
        CenterTr \cite{zhou2020tracking} & 40.3 & 53.8 & 36.9 & 73.0 & 88.8 & 71.1 \\
        PermaTr \cite{tokmakov2021learning} & 47.4 & 65.0 & 43.6 & \textbf{77.4} & \textbf{90.8} & \textbf{77.6} \\
        PolarMOT \cite{kim2022polarmot} & 43.6 & 46.9 & 48.1 & 75.1 & 80.0 & 76.9 \\
        TripletTrack \cite{marinello2022triplettrack} & 50.0 & 46.5 & 57.8 & 73.6 & 84.3 & 74.7 \\
        \midrule
        \multicolumn{7}{l}{\textit{End-to-end}} \\
        \addlinespace[0.2em]
        \rowcolor{green!8}  
        DecoderTracker & \textbf{51.5} & \textbf{56.8} & \textbf{58.4} & 72.2 & 83.2 & 73.8 \\
        \rowcolor{green!12}
        DecoderTracker+ & \textbf{51.1} & \textbf{56.4} & \textbf{58.8} & 72.6 & 82.1 & 73.9 \\
        \bottomrule
    \end{tabular}
\end{table*}

Table \ref{result_KITTI} provides an overview of DecoderTracker's performance on the KITTI benchmark. DecoderTracker achieves competitive results, with HOTA of 70.2 and 82.9 for Car tracking and Pedestrian tracking, respectively. These results position DecoderTracker as a strong contender in MOT on the KITTI.


\begin{figure}
	\centering
	\includegraphics[width=1\textwidth]{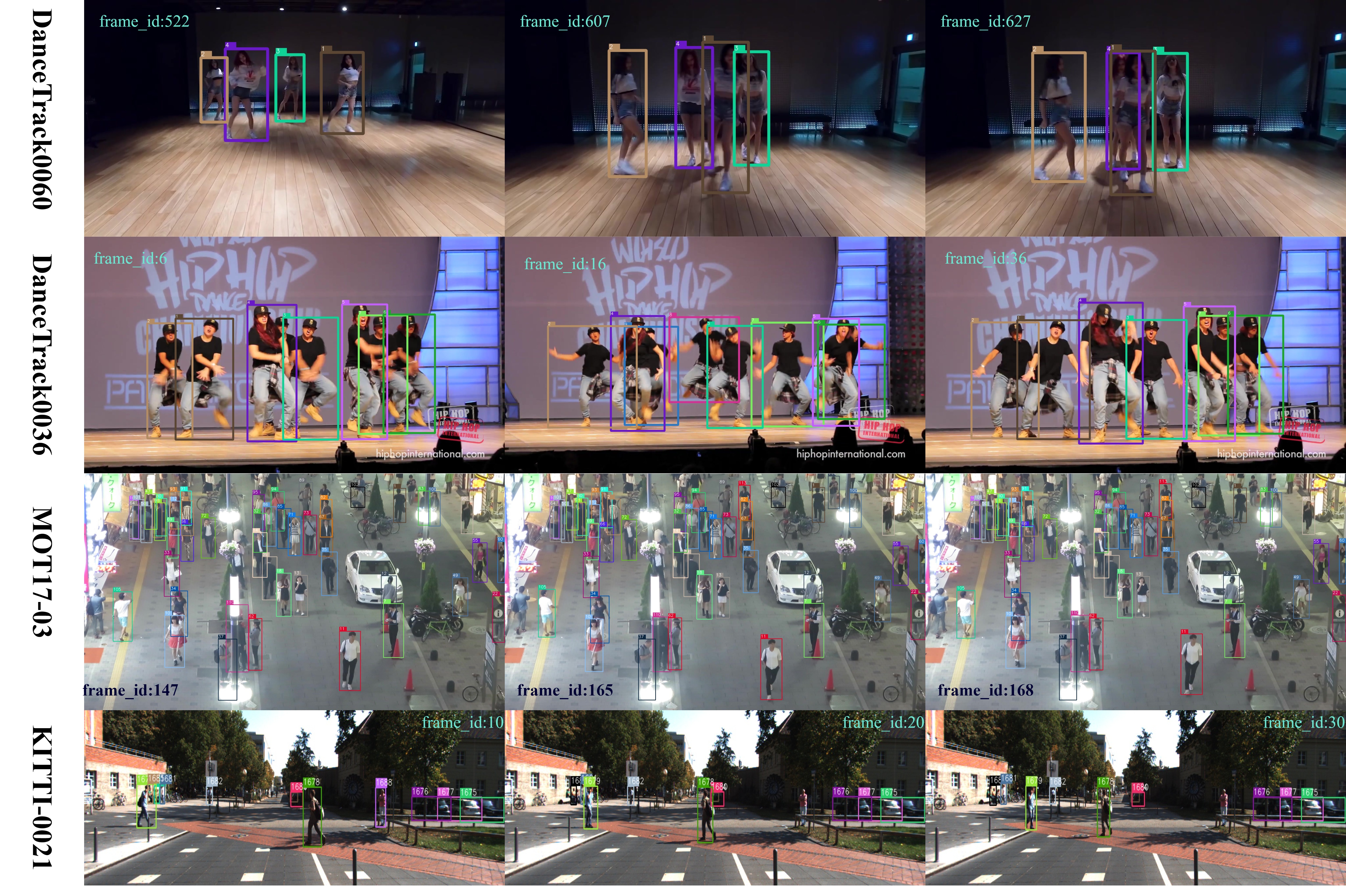}
	\caption{The visualisation results of our proposed DecoderTracker on various datasets show that DecoderTracker performs well both under long time series and short time series, and also has good tracking performance for different kinds of objects. The visualization results of DecoderTracker and DecoderTracker+ are essentially identical, so we won't display the results of D+ here.}
	\label{tracking_result_all}
\end{figure}

\subsection{Ablation Study Of DecoderTracker}

To assess the proposed training strategy and the efficacy of TBSP, ablation experiments were conducted on the DanceTrack benchmark. The training set includes the entire DanceTrack training data, with testing involving 25 sequences from the validation set. Additionally, the impact of two methods for generating query positions—utilizing the sinusoidal functions for encodings from DAB-DETR \cite{liu2022dab} or employing an MLP, on the method's performance was investigated.


\begin{table*}
	\resizebox{\textwidth}{!}{
		\begin{minipage}{.5\textwidth}
			\centering
			\begin{subtable}{\textwidth}
				\centering
				\captionsetup{justification=raggedright, singlelinecheck=false} 
				\caption{Performance of DecoderTracker at different training stages}
				\label{ablation_results_training}
				\setlength{\tabcolsep}{6pt}
				\begin{tabular}{l|ccc}
					\toprule
					Training Stage & IDF1 & HOTA & MOTA \\
					\midrule
					First              & 38.6 & 38.5 & 76.8 \\
					Second             & 44.3 & 44.5 & 84.2 \\
					Final (TBSP)       & 51.7 & 51.9 & 84.9 \\
					Final              & 53.4 & 53.6 & 84.7 \\
					\bottomrule
				\end{tabular}
			\end{subtable}
			
			\begin{subtable}{\textwidth}
				\centering
				\captionsetup{justification=raggedright, singlelinecheck=false} 
				\caption{ The impact of $\delta_{iou}$ on method performance when using TBSP}
				\label{ablation_results_TBSP}
				\setlength{\tabcolsep}{6pt}
				\begin{tabular}{l|ccc}
					\toprule
					$\delta_{iou}$ & IDF1 & HOTA & MOTA \\
					\midrule
					0.3   & 50.1 & 49.7 & 79.1 \\
					0.5   & 51.7 & 51.9 & 84.9 \\
					0.7   & 51.3 & 51.3 & 83.2 \\
					0.9   & 49.8 & 50.0 & 81.3 \\
					\bottomrule
				\end{tabular}
			\end{subtable}
		\end{minipage}%
		\begin{minipage}{.5\textwidth}
			\centering
			\begin{subtable}{\textwidth}
				\centering
				\captionsetup{justification=raggedright, singlelinecheck=false} 
				\caption{
					The presence or absence of TAN during the \textbf{second training stages} can significantly impact the performance of the method.}
				\label{ablation_results_TAN}
				\setlength{\tabcolsep}{6pt}
				\begin{tabular}{l|ccc}
					\toprule
					method & IDF1 & HOTA & MOTA \\
					\midrule
					Second              & 44.4 & 44.5 & 84.2 \\
					Second(w/o)         & 41.9 & 42.1 & 80.8 \\
					Final               & 53.4 & 53.6 & 84.7 \\
					Final(w/o)          & 53.2 & 53.4 & 84.5 \\
					\bottomrule
				\end{tabular}
			\end{subtable}

			\begin{subtable}{\textwidth}
				\centering
				\captionsetup{justification=raggedright, singlelinecheck=false} 
				\caption{Effect of query position generation strategy}
				\label{ablation_results_position}
				\setlength{\tabcolsep}{4pt}
				\begin{tabular}{l|ccc}
					\toprule
					Setting & IDF1 & HOTA & FPS \\
					\midrule
					sin/cos & 85.1 & 53.3 & 18.5 \\
					MLP     & 84.7 & 53.6 & 19.6 \\
					\bottomrule
				\end{tabular}
			\end{subtable}
		\end{minipage}
	}
	\caption{Ablation Studies on Our Proposed DecoderTracker on the DanceTrack validation set.}
	\label{combined_ablation_results}
\end{table*}


The performance of DecoderTracker improves consistently across different training stages, peaking in the final stage. The choice of $\delta_{iou}$ impacts precision and recall trade-offs, with higher values leading to a slight decrease in performance. Position generation strategies, MLP and sinusoidal functions, yield similar performance in terms of detection accuracy, with slight differences in FPS. Overall, the ablation studies demonstrate the effectiveness of DecoderTracker across different training conditions and method configurations, showcasing its adaptability and versatility in MOT field.

It is noteworthy that in Table \ref{ablation_results_TAN}, we demonstrate the impact of the presence or absence of TAN during the second phase of training on the network's tracking performance at different stages. It can be observed that the presence or absence of TAN in the second phase has almost no effect on the final network performance. However, the absence of TAN does impact the tracking performance during the second phase, which also validates the effectiveness of our proposed TBSP strategy in enhancing tracking capabilities.

\subsubsection{More Discusstion About TBSP}

The initial reason for developing the \textcolor{black}{DecoderTracker} method was not due to slow inference speed, but rather the extensive training resources required. Using eight V100 GPUs with pretrained weights, training on an extended MOT17 dataset, which contained approximately 10,000 images, required about 84 hours for 200 epochs. Each epoch took around 0.42 hours, with each GPU processing about 3,000 images per hour.

Fortunately, our DecoderTracker method has addressed this issue. Although its inference speed is not yet real-time, its convergence rate is already satisfactory. Although we have not achieved real-time inference speeds, we have significantly improved the speed of MOTR.

Regarding the TBSP, its original purpose was simply to endow the detection network with basic tracking functionality. However, it also provided some unexpected benefits, such as further reducing training time. Below, we list the results and required iterations if training was conducted directly using TALA, as shown in Table \ref{result_TBSP}.

We believe the effectiveness of the TBSP strategy may be due to its function as a form of weak or self-supervised training for motion features, compared to TALA. Many examples demonstrate that undergoing weakly supervised training before strongly supervised training can enhance performance and speed up convergence. For instance, in the NLP field, large language methods like BERT\cite{devlin2018bert} and GPT\cite{brown2020language} initially train on a vast amount of unlabeled or weakly labeled data to learn basic language patterns. Subsequently, these methods undergo strong supervision fine-tuning for specific tasks like sentiment analysis or question answering. This approach has been empirically proven to significantly boost the performance of LLMs.
\begin{table}
	\begin{center}
		\caption{Comparison of DecoderTracker with and without using TBSP on the DanceTrack val set.}
		\label{result_TBSP}
		\setlength{\tabcolsep}{6pt}
		\begin{tabular}{c|cccc}
			\hline\noalign{\smallskip}
			Setting &IDF1  & HOTA & Epochs & Total Training Time \\
			\hline
			TBSP&  54.2& 53.6 & 27 & 56.7 hours \\
			w/o&  53.5& 53.4 &  40 &   85.2 hours
			\\
			\noalign{\smallskip}
			\hline
		\end{tabular}
	\end{center}
\end{table}
In our field of computer vision, this technique is also quite common, with the most classic example being the SimCLR series\cite{chen2020simple,chen2020big}. Typically, these methods are pretrained on unlabeled data to learn image features, then fine-tuned on strongly supervised tasks such as classification or detection.


Although TBSP did not enhance method performance, it reduced the number of iterations needed for convergence, which might be considered its greatest contribution.

\subsection{Ablation Study Of DecoderTracker+}
As can be seen from the results in Tables \ref{result_dance}, \ref{result_MOT}, and \ref{result_KITTI}, the performance of DecoderTracker and DecoderTracker+ is almost the same. This is not difficult to understand. In Section 3.3, we discussed that empty queries do not affect the network's output results after reasonable mask generation. DecoderTracker+ does not improve the network; instead, it addresses the optimization issues caused by dynamic data. At the same time, to better illustrate the effectiveness of the generated masks, we provided the performance of DecoderTracker+ with and without masks on the DanceTrack val set in Table \ref{result_mask}. The results show that when masks are not used, the inactive queries still have an impact on the final tracking results. Combining this with previous data, both theoretical and practical evidence demonstrates that the use of masks does not affect tracking performance when FSQM is increased after reasonable mask generation.

\begin{table}
	\begin{center}
		\caption{The impact of mask on the performance of DecoderTracker+ on the DanceTrack val set.}
		\label{result_mask}
		\setlength{\tabcolsep}{6pt}
		\begin{tabular}{c|cccc}
			\hline\noalign{\smallskip}
			mask & IDF1&HOTA & MOTA & AssA  \\
			\hline
			\checkmark& 53.6 & 53.4 & 84.9 & 36.0 \\
			w/o &46.2 & 45.8 &  78.3 & 31.7  
			\\
			\noalign{\smallskip}
			\hline
		\end{tabular}
	\end{center}
\end{table}

\subsubsection{Experimental Analysis of FENet}
\begin{figure}
	\centering
	\includegraphics[width=0.5\textwidth]{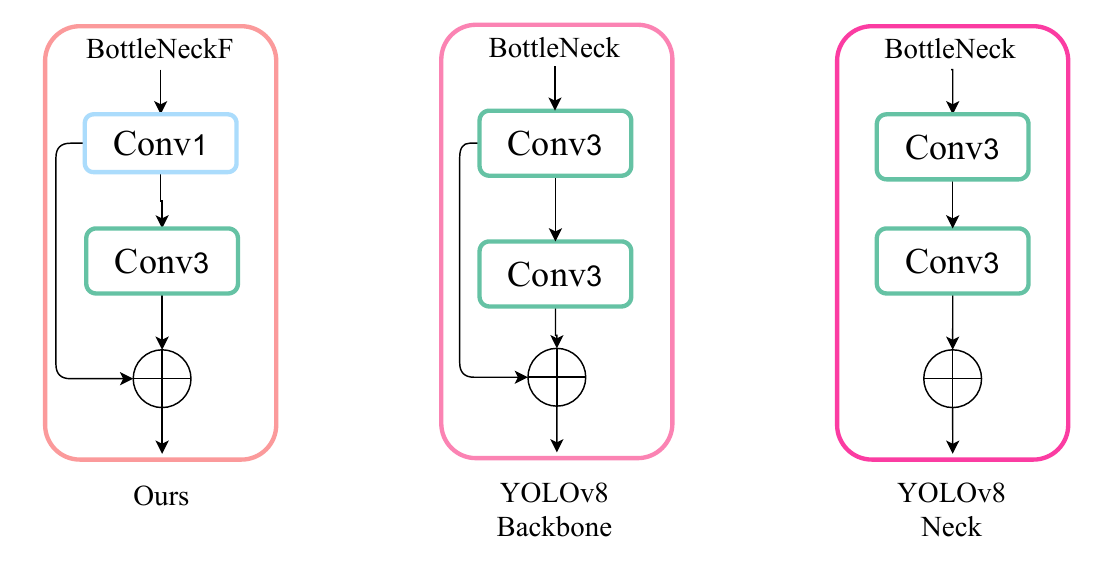}
	\caption{The difference between our proposed method and YOLOv8 in terms of modules. Our modules are more lightweight. Specific method details can refer to the Figure.\ref{detail}. Additionally, it should be noted that there is not much difference between this part of YOLO11 and YOLOv8, so no comparison has been made}
	\label{yolov8_diffenence}
\end{figure}

Furthermore, we have made some modifications to the C2F module in YOLOv8, renaming our modified module to C2M. This change is specifically reflected in the BottleNeckF module. In YOLOv8, there is a BottleNeck module that differs slightly between the neck and backbone; we eliminated this difference and replaced the 3x3 convolution in the middle with a 1x1 convolution to further reduce the computational load. These modifications, grounded in YOLOv8's proven stability and migration performance (with YOLO11's key changes focusing more on the head section, less relevant to our feature extraction adjustments), are illustrated in Figure \ref{yolov8_diffenence}, and their impact on performance is shown in Table \ref{result_YOLO}.

\begin{table}
	\begin{center}
		\caption{The impact of our network and YOLOv8 structure on DecoderTracker and DecoderTracker+ performance on the DanceTrack val set. }
		\label{result_YOLO}
		\setlength{\tabcolsep}{6pt}
		\begin{tabular}{c|ccccc}
			\hline\noalign{\smallskip}
			Setting & IDF1&HOTA & MOTA & AssA & FPS \\
			\hline
			\textit{DecoderTracker:} &&&&&\\
			FENet& 53.8 & 53.6 & 84.7 & 35.9&19.6 \\
			YOLOv8 &51.4 & 52.4 &  84.2 & 36.1  &18.9
			\\
			\hline
			\textit{DecoderTracker+:} &&&&&\\
			FENet& 53.6 & 53.4 & 84.9 & 35.7& 28.2 \\
			YOLOv8 &52.1 & 53.2 &  85.4 & 36.8  & 26.7
			\\
			
			\noalign{\smallskip}
			\hline
		\end{tabular}
	\end{center}
\end{table}

\section{Conclusion and Limitation}
\begin{figure*}
	\centering
	\includegraphics[width=1\textwidth]{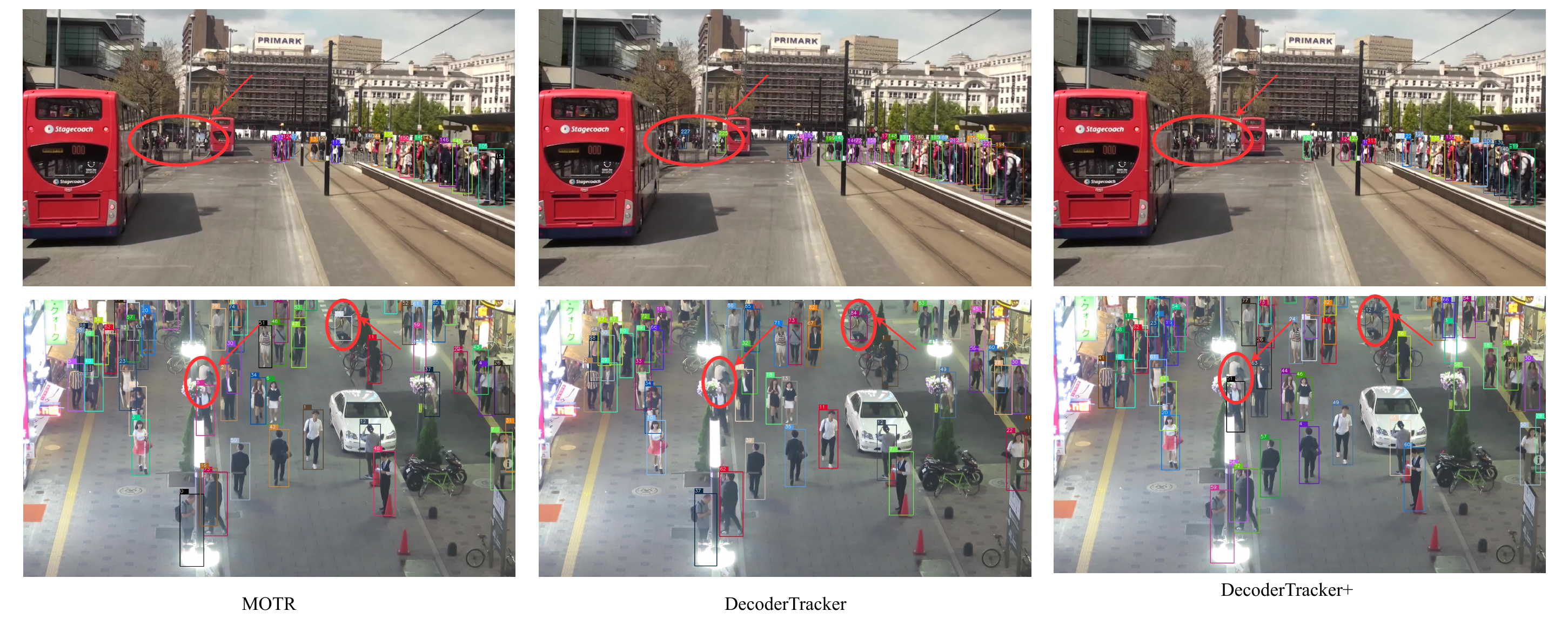}
	\caption{DecoderTracker and MOTR failed in the same scenario. The probable cause may lie in the insufficient scale of the MOT17 benchmark.}
	\label{result}
\end{figure*}


This paper presents DecoderTracker and its enhanced variant DecoderTracker+, two novel end-to-end MOT methods based on a decoder-only architecture, aiming to address the issues of high computational resource consumption, suboptimal inference speed, and prolonged training time in traditional transformer-based end-to-end MOT methods like MOTR. DecoderTracker optimizes the network architecture by removing the redundant encoder and adopting a lightweight Feature Extraction Network (FENet) derived from YOLOv8, which enhances efficiency and inference speed. It also introduces a three-stage training strategy, incorporating TBSP as a weakly supervised training step to accelerate convergence and reduce training time. To tackle optimization challenges from dynamic data, DecoderTracker+ is proposed by integrating FSQM and refining attention layers with an attention masking strategy. FSQM maintains a constant number of tracking queries, enabling better compatibility with modern deep learning frameworks and facilitating deployment, while the masking strategy ensures that inactive queries do not interfere with computations. Experimental results on benchmarks including DanceTrack, MOT17, and KITTI demonstrate that both methods outperform MOTR without additional tricks. DecoderTracker+ achieves an inference speed of 28.8 FPS, which is 2-3 times faster than MOTR, while maintaining competitive performance in metrics such as HOTA, IDF1, and MOTA. This work provides an efficient baseline for end-to-end MOT, laying the foundation for real-time end-to-end MOT methods and offering insights for engineering deployment, with potential future improvements including enhancing temporal modeling and optimizing label assignment strategies .


\color{black}DecoderTracker's primary limitation remains its accuracy gap on MOT17 compared to TBD(e.g., Deep OC-SORT), similar to MOTR's challenge. We attribute this in part to the limited volume of MOT17 training data, which affects end-to-end training effectiveness (Figure \ref{result}). Although our method significantly improves speed and outperforms MOTR, it still trades accuracy for real-time efficiency, a core design choice. Additional challenges include robustness in high-occlusion scenarios (Fig. 10) and domain generalization. Future work will focus on improving occlusion handling, exploring domain adaptation, and refining label assignment strategies to close the accuracy gap. \color{black}

\newpage
\bibliographystyle{elsarticle-num}
\bibliography{natbib}

\end{document}